\documentclass[letterpaper]{article} % DO NOT CHANGE THIS
\usepackage{aaai2026}  % DO NOT CHANGE THIS
\nocopyright
\usepackage{times}  % DO NOT CHANGE THIS
\usepackage{helvet}  % DO NOT CHANGE THIS
\usepackage{courier}  % DO NOT CHANGE THIS
\usepackage[hyphens]{url}  % DO NOT CHANGE THIS
\usepackage{graphicx} % DO NOT CHANGE THIS
\usepackage{natbib}  % DO NOT CHANGE THIS AND DO NOT ADD ANY OPTIONS TO IT
\usepackage{caption} % DO NOT CHANGE THIS AND DO NOT ADD ANY OPTIONS TO IT
\usepackage{algorithm}
\usepackage{algorithmic}
\usepackage{multirow}
\usepackage{subcaption}
\usepackage{mdframed}
\usepackage{booktabs}       % professional-quality tables
\usepackage{bm}
\usepackage{amsmath}
\usepackage{amsfonts}

\usepackage{newfloat}
\usepackage{listings}
\DeclareCaptionStyle{ruled}{labelfont=normalfont,labelsep=colon,strut=off} % DO NOT CHANGE THIS
\usepackage{listings}
\usepackage[T1]{fontenc} % 句読点やクォートの表示安定化
\lstdefinelanguage{json}{
  morestring=[b]",
  morestring=[d]'
  ,literate={`}{{`}}1 % バックスラッシュ等の崩れ回避は必要に応じて
}
\lstdefinestyle{aaai-json}{
  language=json,
  basicstyle=\ttfamily\footnotesize,
  columns=fullflexible,
  keepspaces=true,
  showstringspaces=false,
  breaklines=true,
  breakatwhitespace=true,
  numbers=none,              % ← 行番号を出したいなら 'left'
  numberstyle=\tiny,
  numbersep=5pt,
  xleftmargin=0pt,
  framexleftmargin=0pt,
  frame=single,              % 枠不要なら削除
  captionpos=b,
  aboveskip=4pt,
  belowskip=4pt,
  upquote=true,
  tabsize=2
}

\lstdefinelanguage{prompt}{
  morekeywords={Knowledge,Sentence},
  sensitive=false,
  alsoletter={:,_} % コロンやアンダースコアを単語に含める
}

\lstdefinestyle{aaai-prompt}{
  language=prompt,
  basicstyle=\ttfamily\footnotesize,
  keywordstyle=\bfseries,
  columns=fullflexible,
  keepspaces=true,
  showstringspaces=false,
  breaklines=true,
  breakatwhitespace=true,
  numbers=none,
  frame=single,            % 枠不要なら削除
  captionpos=b,            % キャプションを下に
  xleftmargin=0pt,
  framexleftmargin=0pt,
  aboveskip=4pt,
  belowskip=4pt,
  upquote=true,
  tabsize=2
}

\lstdefinestyle{aaai-list}{
  basicstyle=\ttfamily\footnotesize,
  columns=fullflexible, keepspaces=true,
  showstringspaces=false, breaklines=true, breakatwhitespace=true,
  numbers=none,       % 行番号を出すなら left + numberstyle=\tiny など
  tabsize=2, upquote=true,
  backgroundcolor={}
}

\floatstyle{ruled}
\newfloat{listing}{tb}{lst}{}
\floatname{listing}{Listing}
\title{Concept Unlearning in Large Language Models\\ via Self-Constructed Knowledge Triplets}
\author{
Tomoya Yamashita, Yuuki Yamanaka, Masanori Yamada, Takayuki Miura, Toshiki Shibahara, Tomoharu Iwata
}
\affiliations{
NTT\\
tomoya.yamashita@ntt.com, yuuki.yamanaka@ntt.com, masanori.yamada@ntt.com, tkyk.miura@ntt.com, toshiki.shibahara@ntt.com, tomoharu.iwata@ntt.com
}

\usepackage{bibentry}
\begin{document}

\maketitle

\begin{abstract}
Machine Unlearning (MU) has recently attracted considerable attention as a solution to privacy and copyright issues in large language models (LLMs).
Existing MU methods aim to remove specific target sentences from an LLM while minimizing damage to unrelated knowledge.
However, these approaches require explicit target sentences and do not support removing broader concepts, such as persons or events.
To address this limitation, we introduce Concept Unlearning (CU) as a new requirement for LLM unlearning.
We leverage knowledge graphs to represent the LLM's internal knowledge and define CU as removing the forgetting target nodes and associated edges.
This graph-based formulation enables a more intuitive unlearning and facilitates the design of more effective methods.
We propose a novel method that prompts the LLM to generate knowledge triplets and explanatory sentences about the forgetting target and applies the unlearning process to these representations.
Our approach enables more precise and comprehensive concept removal by aligning the unlearning process with the LLM's internal knowledge representations.
Experiments on real-world and synthetic datasets demonstrate that our method effectively achieves concept-level unlearning while preserving unrelated knowledge.
\end{abstract}

\begin{figure*}[tb]
\centering
\includegraphics[width=0.9\linewidth]{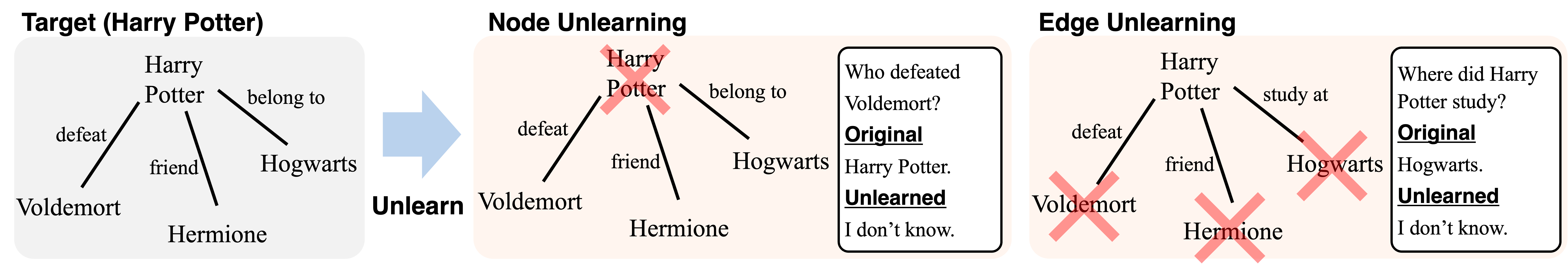}
\caption{\textbf{Overview of Concept Unlearning (CU).}
The left side shows the knowledge graph of the forgetting target.
The middle side shows the first sub-requirement of CU, \textbf{Node Unlearning}, where the LLM should not output the target entity (e.g., ``Harry Potter'') even when prompted with associated facts.
The right side shows the second sub-requirement, \textbf{Edge Unlearning}, where the LLM should not output the attributions (e.g., ``Hogwarts'') even when the target entity is mentioned.
}
\label{CU_example}
\end{figure*}

\section{Introduction}
Large language models (LLMs) have become core components of modern infrastructure, enabling applications in text generation, data analysis, and specialized areas such as medicine, law, and programming~\citep{gpt3,jiang2023mistral,grattafiori2024llama3herdmodels,yang2024qwen2}.
However, training data (e.g., Web corpora) often contain sensitive or undesirable information, including Personally Identifiable Information.
Existing studies have shown that LLMs trained on such data may memorize and unintentionally expose private information, posing serious privacy and security risks~\citep{Carlini2020ExtractingTD,DBLP:conf/iclr/CarliniIJLTZ23}.
Moreover, data protection regulations such as the General Data Protection Regulation (GDPR)~\citep{10.5555/3152676} require LLM providers to respond promptly to information deletion requests. 
Addressing these issues is essential for safe, trustworthy, and compliant LLM deployment. \par
Machine Unlearning (MU) has received significant attention as a potential solution to the above problems~\citep{10.1145/3603620}.
MU aims to remove the influence of arbitrary training data from a trained model.
The standard MU framework typically involves the following steps: (1) identify the target data points in the training dataset, and (2) apply the unlearning process to those data points.
Existing MU research primarily focuses on improving forgetting performance in step (2), ensuring the model effectively erases the influence of the target data points without collateral damage~\citep{bourtoule2021machine,ginart2019making,neel2021descent,yao2024large}.
However, in the context of LLMs, deletion requests correspond not necessarily to specific data points (e.g., sentences) but to broader concepts such as persons or events.
In real-world scenarios, it may be necessary to remove knowledge about individuals involved in privacy issues or about incidents that become sensitive.
In such cases, step (1) of the MU framework, identifying the target data points, becomes particularly challenging because it is often unclear which training examples are responsible for a given concept. 
Unlike discrete sentences, concepts are typically distributed across many data points in complex and non-obvious ways.
As a result, these scenarios call for concept-level unlearning without the explicitly identifiable training data.
Furthermore, the original training corpus is often no longer accessible, and the corresponding sentences about the forgetting target cannot be located and removed.
This practical barrier also challenges existing MU techniques.
Current MU research has not sufficiently addressed these issues, leaving the problem of reliably unlearning concept-level knowledge unresolved. \par
To solve the above problems, we propose \textbf{Concept Unlearning (CU)}, a new requirement for LLM unlearning.
As the name suggests, CU targets removing specific ``concepts'' rather than the sentences or tokens.
We formalize CU with a knowledge graph, where nodes represent entities (e.g., persons, events), and edges represent their relationships.
CU defines the forgetting target concept as a node and its connections within the LLM's internal knowledge graph.
In other words, CU requires the LLM to suppress the output of the target entity (\textbf{Node Unlearning}) and its associated attributions (\textbf{Edge Unlearning}) under relevant prompts.
We show the overview of CU in Fig.~\ref{CU_example}.
CU provides a precise formulation of concept-level forgetting without explicit target examples, thereby improving the interpretability and feasibility of the unlearning process.
Note that, in this paper, we restrict the term ``concept'' to the knowledge expressed in the triplet form, specifically as the set of triplets that include the name of the forgetting target.
Extending unlearning to other types of knowledge, such as procedural knowledge (e.g., how to make a bomb), is left for future work. \par
We propose a novel method to realize CU. 
Two key components characterize our approach.
First, we leverage the target LLM to generate knowledge triplets about the forgetting target; these structured facts become explicit unlearning targets that efficiently erase the associated attributions.
Second, we collect self-generated explanatory sentences about the forgetting target and apply the unlearning process to these sentences.
This dual-target strategy enables our method to focus on LLM's internal knowledge, leading to more complete and effective unlearning.
We evaluate our method using the real-world original \textsc{Wiki-Fact} dataset and the synthetic \textsc{TOFU} dataset~\citep{maini2024tofu}.
We test whether the LLM outputs the target entity or its associated attributions under relevant prompts.
Experimental results demonstrate that our method successfully removes concept-level knowledge while preserving unrelated knowledge.
We also conduct ablation studies to quantify the contribution of each component and identify factors that influence unlearning performance.
Our contributions are as follows:
\begin{itemize}
    \item We introduce Concept Unlearning as a new requirement for LLM unlearning, which aims to remove specific ``concepts'' rather than the sentences or tokens.
    \item We propose a novel self-supervised unlearning method that leverages the target LLM to unlearn its own knowledge, enabling effective and selective forgetting.
    \item We empirically show that our method achieves complete forgetting while preserving unrelated knowledge and general utility on real-world and synthetic datasets.
\end{itemize}

\section{Related Works}
\subsection{Machine Unlearning}
Machine Unlearning (MU) aims to remove the effects of arbitrary training data from a model~\citep{bourtoule2021machine,ginart2019making,neel2021descent}.
Recently, MU has been studied in the context of LLMs~\citep{yao2024large,zhang2024negative,pmlr-v235-li24bc,pmlr-v235-pawelczyk24a}.
\citet{yao2024large} suppressed harmful or copyrighted behaviors using negative examples.  
\citet{zhang2024negative} proposed negative preference optimization (NPO), improving efficiency and stability over naive unlearning processes such as gradient ascent (GA). 
\citet{pmlr-v235-li24bc} introduced representation misdirection for unlearning to suppress hazardous knowledge.  
\citet{pmlr-v235-pawelczyk24a} developed in-context unlearning (ICU), which intervenes at inference time without parameter updates. 
Although some works aim to forget factual knowledge rather than individual data points~\citep{maini2024tofu,jang-etal-2023-knowledge,eldan2023whosharrypotterapproximate,jin2024rwku}, they still operate at the sentence level, limited to isolated QA-style answers without modeling each factual knowledge as a structured representation.
In contrast, our work explicitly defines CU, which targets concept-level knowledge via knowledge graphs. 
This graph-based definition allows systematic removal and enables interpretable evaluation consistent with the CU formulation.

\subsection{Knowledge Graphs for LLMs}
Knowledge graphs are sets of structured factual knowledge in the triplet form $(\bm{s}, \bm{r}, \bm{o})$, where $\bm{s}$ denotes a subject, $\bm{r}$ a relation, and $\bm{o}$ an object~\citep{8047276}.  
For example, the fact ``Donald John Trump was born in New York'' can be represented as ($\bm{s}=$ ``Donald John Trump'', $\bm{r}=$ ``place of birth'', $\bm{o}=$ ``New York'').
Knowledge graphs have been utilized to probe and edit the factual knowledge embedded within LLMs.  
In the following subsections, we summarize key developments in these areas.
Note that we use the term ``knowledge triplets'' to refer to individual $(\bm{s}, \bm{r}, \bm{o})$ facts, and ``knowledge graphs'' to denote a collection of such triplets.
\subsubsection{LLM Probing}
LLM probing uses knowledge graphs to assess the factual knowledge stored in LLMs.
\citet{Petroni2019LanguageMA} introduced \textsc{LAMA}, converting triplet facts into cloze-style prompts to probe LLM's knowledge.
\citet{jiang-etal-2020-know} extended \textsc{LAMA} with \textsc{LPAQA} by enriching paraphrased templates.
\citet{shin-etal-2020-autoprompt} proposed \textsc{AutoPrompt}, a gradient-based method to automatically generate prompts.
\citet{meng-etal-2022-rewire} focused on domain-specific knowledge with \textsc{MedLAMA}, and \citet{luo2023systematic} developed a systematic framework for generating probing questions from triplets.
These studies demonstrate that probing LLMs via knowledge graphs provides a powerful framework for extracting and analyzing their internal knowledge.
Inspired by these approaches, we utilize knowledge graphs to achieve concept-level unlearning systematically.

\subsubsection{Knowledge Edit}
Knowledge Edit aims to modify the factual knowledge stored in LLMs by updating their model parameters~\citep{dai2022knowledge,meng2022locating,meng2023massediting,mitchell2022fast,DBLP:conf/iclr/FangJWMSW0C25}.
They focus on editing LLM's internal triplets, replacing an original object $\bm{o}$ with a new value $\bm{o}'$.
\citet{dai2022knowledge} introduced knowledge neurons, enabling direct editing of the LLM's internal knowledge.
\citet{meng2022locating} presented ROME, which updates knowledge via localized rank-one edits.
\citet{meng2023massediting} introduced MEMIT, which efficiently edits thousands of pieces of knowledge in LLMs.
\citet{mitchell2022fast} designed MEND using small auxiliary networks to apply fast, low-rank edits. 
\citet{DBLP:conf/iclr/FangJWMSW0C25} introduced a null-space projection to mitigate interference during editing.
While Knowledge Edit may be applied to remove specific facts, it requires explicit specification of the target knowledge, and thus shares the same limitations as existing MU methods in achieving concept-level forgetting.

\section{Concept Unlearning}
\label{cu_section}
CU is a new MU requirement for removing specific concepts from an LLM.
We define a ``concept'' as an entity (e.g., a person or an event) and its directly associated attributions, represented with LLM's internal knowledge triplets.
Specifically, we define the target entity as $\bm{e}^\mathrm{(t)}$ and define concept as the set of knowledge triplets $U^\mathrm{(t)}$ in which $\bm{e}^\mathrm{(t)}$ appears either as the subject or the object (i.e., $\bm{s} = \bm{e}^\mathrm{(t)}$ or $\bm{o} = \bm{e}^\mathrm{(t)}$).
We thus characterize a concept as a local semantic neighborhood in the knowledge graph, centered at the target entity and composed of its factual relations. 
An overview of CU is shown in Fig.~\ref{CU_example}, where the target is ``Harry Potter''.
Intuitively, CU aims to ensure that the LLM does not regenerate the target-related knowledge in its output unless such knowledge is already explicitly provided in the prompt.
We formally define this CU requirement as:
\begin{align}
    \forall \bm{x}\in \mathcal{V}^*, \forall (\bm{s}, \bm{r}, \bm{o})\in U^\mathrm{(t)}, (\bm{s}, \bm{r}, \bm{o})\not\in g(\bm{x}) \nonumber \\
    \Rightarrow (\bm{s}, \bm{r}, \bm{o}) \not\in g\left(\mathrm{Concat}\{\bm{x}, \mathrm{LLM}(\bm{x})\}\right),
    \label{cu_requirement}
\end{align}
where $g$ is a function that extracts knowledge triplets from a sentence~\footnote{For example, GPT-4o with an appropriate prompt can be used as a knowledge extraction function.}, $\bm{x} \in \mathcal{V}^*$ is the input token sequence (where $\mathcal{V}$ denotes the vocabulary), $\mathrm{LLM}(\bm{x})$ is the output of the LLM, and $\mathrm{Concat}\{\bm{x}, \mathrm{LLM}(\bm{x})\}$ denotes the concatenation of the input and output sequences into a single string.
For example, suppose that $\bm{e}^\mathrm{(t)}$ is ``Harry Potter'', and an input $\bm{x}=$ ``Who defeated Voldemort?''.
If the LLM outputs ``Harry Potter.'', the concatenation of the input and output contains the triplet (``Harry Potter'', ``defeated'', ``Voldemort''), which is in $U^\mathrm{(t)}$.
Therefore, the condition in Eq.~\ref{cu_requirement} is violated. \par
While CU is formally defined as removing concept-level knowledge, verifying this condition under arbitrary inputs is impractical.
In this paper, we introduce two sub-requirements: Node Unlearning and Edge Unlearning, which represent stricter conditions in which parts of the target triplets are provided.
Although they do not capture all possible cases, they provide a practical approximation. 
\par
\noindent{\textbf{Node Unlearning (NU): Do not output the target entity.}}
NU requires that the LLM not output the target entity when given a relevant prompt.
Formally, if a prompt contains some pairs of the relation $\bm{r}$ and the attribution $\bm{o}$ in $U^\mathrm{(t)}$ without the target $\bm{e}^\mathrm{(t)}$, the LLM should not output $\bm{e}^\mathrm{(t)}$.
An example of NU is illustrated on the middle side of Fig.~\ref{CU_example}, where the LLM should not output ``Harry Potter'' even if (-, ``defeat'', ``Voldemort'') is provided.
NU corresponds to removing the target entity node in the LLM's internal knowledge graph. \par
\noindent{\textbf{Edge Unlearning (EU): Do not output the associated attributions.}}
EU requires that the LLM not output associated attributions when given a relevant prompt.
Formally, if a prompt provides the target entity $\bm{e}^\mathrm{(t)}$ and the relation $\bm{r}$ without the attribution $\bm{o}$, the LLM should not output the correct $\bm{o}$.
An example of EU is illustrated on the right side of Fig.~\ref{CU_example}, where the LLM should not output ``Hogwarts'' even if (``Harry Potter'', ``study at'', -) is provided.
EU corresponds to removing the edges between the target entity node and its attribution nodes in the LLM's internal knowledge graph.  \par
By formalizing concept unlearning via a triplet-based structure, we provide a concrete and testable requirement that connects abstract deletion requests to concrete model behavior. 
This enables more precise evaluation, supports the design of targeted probes, and encourages more interpretable and compliant unlearning methods. \par

\section{Proposal}
\label{proposal}
We propose a novel unlearning method that effectively achieves CU while preserving the LLM's overall capabilities.
Our approach has two key losses.
\textbf{(1) Triplet-based loss:} 
This loss suppresses attributions related to the forgetting target using self-generated knowledge triplets, primarily contributing to EU.
\textbf{(2) Sentence-based loss:} 
This loss targets explanatory sentences about the forgetting target, suppressing entity-specific knowledge, and primarily contributes to comprehensive CU by reducing the LLM’s ability to recall the target concept when prompted with related descriptions.
As shown in Algorithm~\ref{proposal_alg}, these two losses are optimized alternately.
We describe each loss in detail below.

\begin{figure}[tb]
\begin{algorithm}[H]
\caption{Our proposed algorithm}
\label{proposal_alg}
\begin{algorithmic}[1]
\REQUIRE Forgetting target: $\bm{e}^\mathrm{(t)}$, Pre-unlearned LLM: $\bm{\theta}_\mathrm{pre}$, Learning rate: $\lambda$, Epochs: $T$
\STATE $\bm{X}_\mathrm{sent} \leftarrow \mathrm{GET\_SENT}_{\bm{\theta}_\mathrm{pre}}(\bm{e}^\mathrm{(t)})$
\FOR{$t$ in $T$}
\STATE $\bm{X}_\mathrm{ent}, \bm{X}_\mathrm{attr}\leftarrow \mathrm{GET\_ATTR}_{\bm{\theta}, \bm{\theta}_\mathrm{pre}}(\bm{e}^\mathrm{(t)})$
\STATE $\bm{\theta}\leftarrow \bm{\theta}-\lambda \nabla_{\bm{\theta}}\mathcal{L}_1(\bm{X}_\mathrm{ent}, \bm{X}_\mathrm{attr})$
\STATE $\bm{\theta}\leftarrow \bm{\theta}-\lambda \nabla_{\bm{\theta}}\mathcal{L}_2(\bm{X}_\mathrm{sent})$
\ENDFOR
\RETURN Unlearned LLM $\bm{\theta}$
\end{algorithmic}
\end{algorithm}
\end{figure}

\subsection{Triplet-based loss: Unlearning with Self-Generated Triplets}
Achieving CU requires suppressing both the target entity and its attributions, corresponding to NU and EU.
Among these, EU presents a distinct challenge due to attributional knowledge's diverse and distributed nature.
While the target entity corresponds to a single node in a knowledge graph, its associated attributions are distributed across multiple relational types (e.g., ``occupation'', ``place of birth'', ``citizenship'', etc.).
We hypothesize that this relational diversity makes it challenging to identify all associated attributions.
To address this challenge, we introduce the \textsc{Get\_Attr} function, which efficiently extracts attributions via LLM's internal knowledge graphs.
The \textsc{Get\_Attr} function prompts the current LLM (i.e., the LLM being unlearned) to generate knowledge triplets related to the forgetting target, such as (``Harry Potter'', ``defeat'', ``Voldemort''), using queries like ``Tell me about \{$\bm{e}^\mathrm{(t)}$\} in knowledge triplet format.''.
Then, we verify each generated triplet with the pre-unlearned LLM $\bm{\theta}_\mathrm{pre}$ and retain only those that the LLM deems valid.
Finally, each validated triplet is converted into a sentence and split into two parts: $\bm{X}_{\mathrm{ent}}$, containing the context up to the entity name, and $\bm{X}_{\mathrm{attr}}$, containing the attribution (e.g., $\bm{X}_{\mathrm{ent}}=$ ``Harry Potter'', $\bm{X}_{\mathrm{attr}}=$ ``defeated Voldemort.'').
This split allows the unlearning process to focus on the attributional tokens.
We use these split sentences and define the triplet-based loss to be minimized under the GA framework:
\begin{align}
    \mathcal{L}_1^\mathrm{GA}(\bm{X}_\mathrm{ent}, \bm{X}_\mathrm{attr}) = \log p_{\bm{\theta}}(\bm{X}_{\mathrm{attr}}|\bm{X}_{\mathrm{ent}}), 
    \label{l_1}
\end{align}
where $\bm{\theta}$ denotes the current LLM and $p_{\bm{\theta}}$ denotes the output distribution.
Minimizing this loss suppresses the LLM's ability to recall attributions when prompted with entity-level context (e.g., ``Who did Harry Potter defeat?'').
Notably, our method is orthogonal to the choice of unlearning algorithm.
We adopt NPO~\citep{zhang2024negative} as our alternative variant:
\begin{align}
    &\mathcal{L}_1^\mathrm{NPO}(\bm{X}_\mathrm{ent}, \bm{X}_\mathrm{attr}) \nonumber \\
    &= \frac{2}{\beta}\log \left(1+\left(\frac{p_{\bm{\theta}}(\bm{X}_{\mathrm{attr}}|\bm{X}_\mathrm{ent})}{p_{\bm{\theta}_\mathrm{pre}}(\bm{X}_{\mathrm{attr}}|\bm{X}_\mathrm{ent})}\right)^\beta\right),
    \label{l_1_npo}
\end{align}
where $\bm{\theta}_\mathrm{pre}$ denotes the pre-unlearned LLM and $\beta$ denotes the inverse temperature hyperparameter.
Importantly, we call \textsc{Get\_Attr} function in each iteration and update $\bm{X}_\mathrm{ent}$ and $\bm{X}_\mathrm{attr}$ throughout the unlearning process, rather than fixing them at the beginning.
This iterative update enables the current LLM to suppress residual attributions, thereby continuously improving coverage and adaptability.
The algorithm of \textsc{Get\_Attr} is provided in Appendix~\ref{get_attr}.

\subsection{Sentence-based loss: Unlearning with Self-Generated Sentences}
We use explanatory sentences generated by the LLM about the forgetting target.
These sentences typically contain multiple attributions and provide rich context.
Unlearning such sentences suppresses general knowledge about the forgetting target and reduces the LLM’s ability to recall the target concept, thereby contributing to comprehensive CU.
To facilitate this, we introduce the \textsc{Get\_Sent} function, which efficiently extracts explanatory sentences about the forgetting target from the pre-unlearned LLM.
The \textsc{Get\_Sent} function prompts the LLM with ``Tell me about \{$\bm{e}^\mathrm{(t)}$\}.'' and obtains a set of sentences $\bm{X}_\mathrm{sent}$, such as ``Harry Potter is a fictional wizard and studied at Hogwarts$\cdots$''.
We use these sentences to calculate the sentence-based loss under the GA framework:
\begin{align}
    \mathcal{L}_2^\mathrm{GA}(\bm{X}_\mathrm{sent}) = \log p_{\bm{\theta}}(\bm{X}_\mathrm{sent}).
    \label{l_2}
\end{align}
By minimizing this loss, we directly lower the likelihood that the LLM will reproduce sentences describing the target entity, thereby suppressing its entity-specific knowledge and enhancing CU in rich context settings.
We also present an NPO variant of the sentence-based loss as:
\begin{align}
    \mathcal{L}_2^\mathrm{NPO}(\bm{X}_\mathrm{sent}) = \frac{2}{\beta}\log \left(1+\left(\frac{p_{\bm{\theta}}(\bm{X}_{\mathrm{sent}})}{p_{\bm{\theta}_\mathrm{pre}}(\bm{X}_{\mathrm{sent}})}\right)^\beta\right).
    \label{l_2_npo}
\end{align}
To comprehensively obtain the forgetting target's explanations, we leverage the pre-unlearned LLM to generate diverse outputs by iteratively asking for explanations.
Note that we call \textsc{Get\_Sent} once at the beginning and use the generated sentences throughout the process.
These fixed sentences ensure optimization stability throughout the unlearning process. 
Furthermore, $\bm{X}_\mathrm{sent}$ is generated via greedy decoding, ensuring high-confident sentences.

\begin{figure*}[tb]
\centering
\begin{minipage}{0.48\textwidth}
\centering
\includegraphics[width=0.75\textwidth]{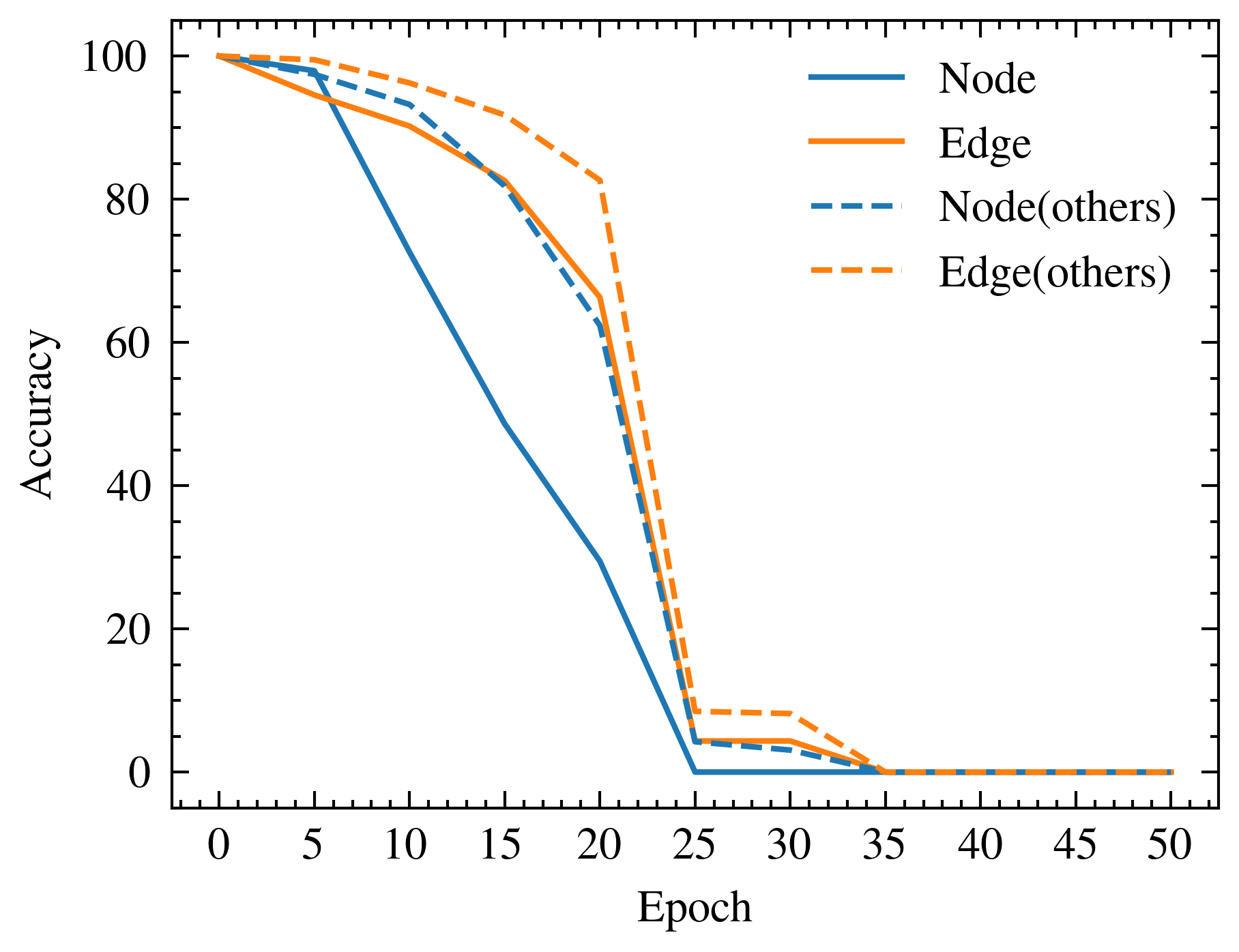}
\subcaption{\textbf{Results of GA.}}
\label{CU_evolution_GA}
\end{minipage}
\begin{minipage}{0.48\textwidth}
\centering
\includegraphics[width=0.75\textwidth]{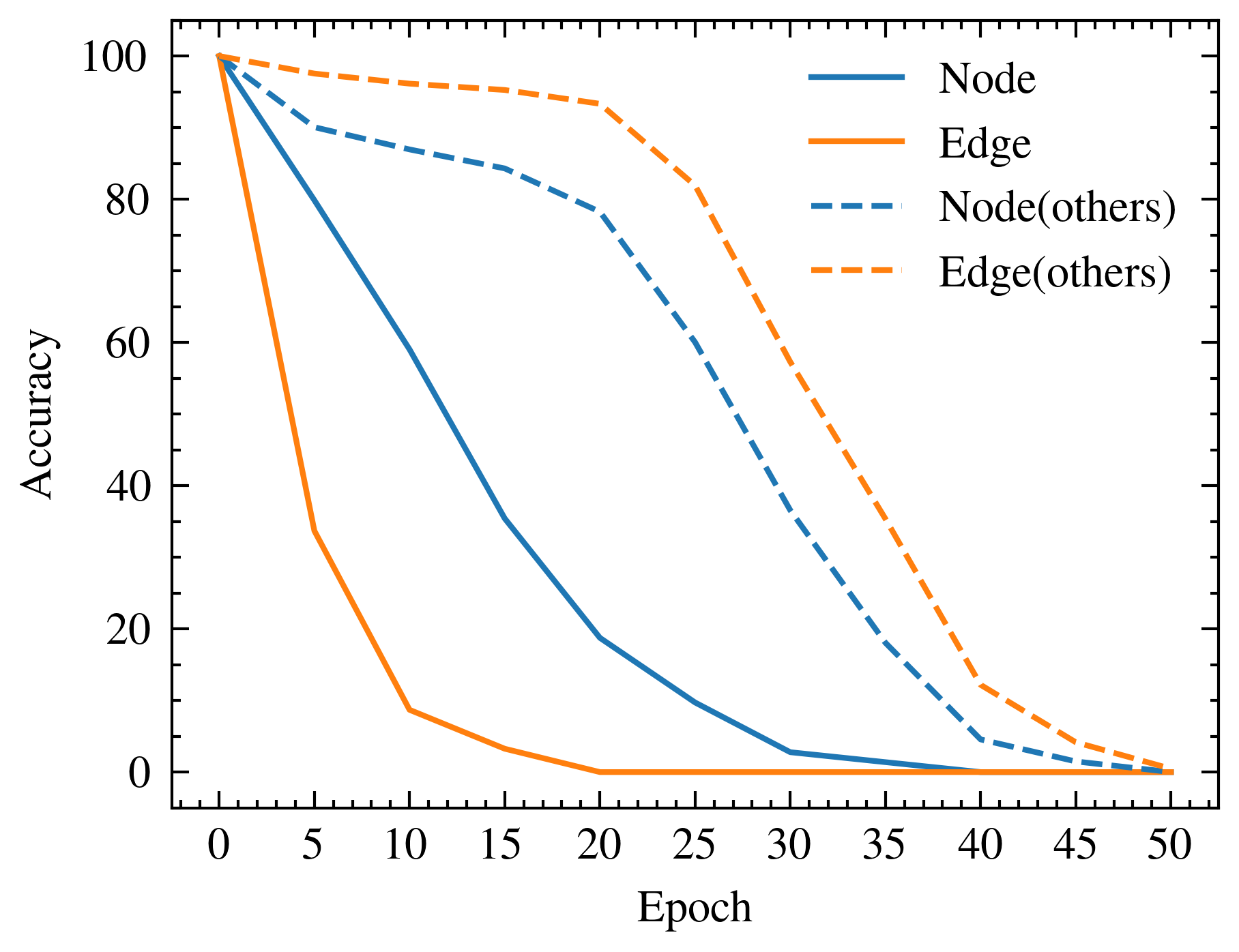}
\subcaption{\textbf{Results of Ours(+GA).}}
\label{CU_evolution_ours}
\end{minipage}
\caption{\textbf{Results on Mistral-7B.}
Blue and orange lines indicate NodeAcc and EdgeAcc, respectively.
Solid lines correspond to target entities, while dashed lines indicate non-target entities.
}
\label{CU_evolution}
\end{figure*}

\section{Experiments}
\label{evaluation}
This section evaluates our method against existing baselines on both real-world and synthetic datasets.
Furthermore, we conduct an ablation study to better understand each loss's contribution in our method, and an additional study to identify the factors influencing forgetting performance.
\subsection{Experimental Setup}
\noindent{\textbf{Datasets:}}
\label{setup}
We conduct experiments on two datasets: the real-world original \textsc{Wiki-Fact} dataset and the synthetic \textsc{TOFU} dataset~\citep{maini2024tofu}.
\textsc{Wiki-Fact} contains real-world human entities, each with a whole Wikipedia explanatory corpus and attributional knowledge from Wikidata.
\textsc{TOFU}, in contrast, consists of 200 fictional human entities. 
Each entity is associated with 20 question–answer sentence pairs.
For each entity in \textsc{TOFU}, we divide the 20 answer sentences into five groups of four answers and combine each group's answers to obtain a separate explanatory corpus. 
This results in five explanatory corpora per entity.
We then use GPT-4 to extract attributional knowledge (i.e., knowledge triplets) from each explanatory corpus. \par
\noindent{\textbf{Metrics:}}
We use two CU metrics, NodeAcc and EdgeAcc. 
NodeAcc measures whether the LLM outputs the target entity when given relevant prompts: we mask the target entity $\bm{e}^\mathrm{(t)}$ in the explanatory corpus and ask the LLM to predict the masked token.
EdgeAcc measures whether the LLM outputs correct attributions about the forgetting target when given relevant prompts. 
To create such prompts, we convert knowledge triplets involving the target entity into natural language sentences and mask the target attribution~\footnote{To ensure realistic and diverse prompts, we leverage the \textsc{ParaRel} dataset~\citep{elazar-etal-2021-measuring}, which provides a set of manually curated paraphrased templates for expressing factual relations in cloze-style formats.}.
These two metrics correspond directly to NU and EU.
We evaluate these metrics for target and non-target entities to assess how well our method forgets specific concepts while preserving unrelated knowledge.
In addition, we evaluate our method on seven utility benchmarks~\citep{zellers-etal-2019-hellaswag,kazemi-etal-2023-lambada,sakaguchi2021winogrande,copa,Clark2018ThinkYH,jin-etal-2019-pubmedqa}~\footnote{Due to space limitations, we report the average performance across seven benchmarks in the main paper, while detailed results for each benchmark are provided in Appendix~\ref{gen_results}}.
Note that, for the CU metrics (NodeAcc and EdgeAcc), we restrict the evaluation to samples that the pre-unlearned LLM correctly answered, ensuring that we measure the extent to which memorized knowledge is forgotten.  
Under this setting, the pre-unlearned LLM achieves 100\% accuracy by definition.  
In contrast, we do not apply such normalization to the general utility score to assess the LLM’s overall performance.
\par
\noindent{\textbf{Models and baselines:}}
We evaluate our method on Mistral-7B Instruction~\citep{jiang2023mistral} and Llama3.1-8B Instruction~\citep{grattafiori2024llama3herdmodels}.
As baselines, we compare against ICU~\citep{pmlr-v235-pawelczyk24a}, GA, and NPO~\citep{zhang2024negative}.
GA and NPO are applied to the explanatory corpus in the datasets.
We conduct a grid search over the number of epochs (step size: 5, max: 50) and report the best result per forgetting target.
All experiments were conducted on two NVIDIA H100 GPUs (80GB each).
Further details on the experimental setup and datasets are provided in Appendix~\ref{eval_details}.

\subsection{Evaluation on Real-World Knowledge}
We apply the unlearning process to each of the 30 entities in the \textsc{Wiki-Fact} and evaluate NU, EU, and general utility metrics. 
The results in Table~\ref{main_results} show that both variants of our method achieve complete unlearning while preserving non-target accuracy.
Notably, our methods attain higher non-target accuracy and better general utility among the baselines that achieve complete unlearning (i.e., NodeAcc = 0.0\% and EdgeAcc = 0.0\%).  
As a reference, the pre-unlearned LLMs achieve general utility scores of 62.2\% for Mistral and 57.6\% for Llama.  
These results indicate a practical advantage of our method: it removes the target knowledge without significantly damaging non-target facts.
\par
Figure~\ref{CU_evolution} illustrates the evolution of CU metrics during the GA-based unlearning process.
Figure~\ref{CU_evolution_GA} shows the results of naive GA applied to the \textsc{Wiki-Fact} corpus.
It reveals that EdgeAcc for the forgetting targets decays only gradually, showing that GA on the \textsc{Wiki-Fact} corpus has difficulty suppressing attributional knowledge.
In addition, the accuracy for both targets and non-targets declines simultaneously as unlearning progresses, suggesting a lack of selectivity. 
In contrast, Fig.~\ref{CU_evolution_ours} demonstrates that Ours (+GA) rapidly reduces EdgeAcc for the forgetting targets, and the accuracy degradation for non-targets is modest.
These results indicate that our method suppresses attributions more effectively and improves selectivity.
This improvement stems from two key features of our method.
First, our triplet-based loss explicitly suppresses attributions associated with the forgetting target.
Second, both losses focus on self-generated content, which is closely aligned with the LLM's internal knowledge representation.
These mechanisms focus on the target concept and achieve more selective forgetting.
We also report the additional results for other LLMs in Appendix~\ref{other_llms}.

\begin{table*}[tb]
\small
    \centering
    \caption{\textbf{Main results on the \textsc{Wiki-Fact} and \textsc{TOFU} datasets.}
Each table reports the unlearning performance on two different LLMs (Mistral-7B and Llama3.1-8B).
We report accuracy on the forgetting targets (Node $\downarrow$, Edge $\downarrow$), non-target knowledge (Node (others) $\uparrow$, Edge (others) $\uparrow$), and general utility (General $\uparrow$).
The best value in each column is shown in bold.}
    \begin{subtable}[tb]{\linewidth}
    \centering
    \caption{Real-world knowledge (\textsc{Wiki-Fact}).}
    \begin{tabular}{llccccc}
        \toprule
        LLM & Method & Node $\downarrow$ & Edge $\downarrow$ & Node (others) $\uparrow$ & Edge (others) $\uparrow$ & General $\uparrow$ \\
        \midrule
        \multirow{5}{*}{Mistral-7B}
        & ICU & 22.6 & 80.4 & 33.0 & 84.4 & 59.8 \\
        & GA & \textbf{0.0} & \textbf{0.0} & 9.3 & 12.4 & 45.0 \\
        & NPO & \textbf{0.0} & 1.1 & 11.2 & 31.4 & 57.1 \\
        & Ours (+GA) & \textbf{0.0} & \textbf{0.0} & \textbf{69.8} & 84.0 & 59.1 \\
        & Ours (+NPO) & \textbf{0.0} & \textbf{0.0} & 68.9 & \textbf{84.5} & \textbf{59.9} \\
        \midrule
        \multirow{5}{*}{Llama3.1-8B}
        & ICU & 33.5 & 86.1 & 31.0 & \textbf{87.1} & \textbf{58.4} \\
        & GA & \textbf{0.0} & \textbf{0.0} & 11.6 & 20.6 & 43.0 \\
        & NPO & 3.7 & 19.8 & 18.7 & 61.1 & 52.7 \\
        & Ours (+GA) & \textbf{0.0} & \textbf{0.0} & 49.3 & 70.3 & 55.1 \\
        & Ours (+NPO) & \textbf{0.0} & \textbf{0.0} & \textbf{50.9} & 76.8 & 55.4  \\
        \bottomrule %\vspace{0.1mm} %三浦加筆
    \end{tabular}
    \label{main_results}
    \end{subtable}
    \begin{subtable}[tb]{\linewidth}
    \centering
    \caption{Fictional knowledge (\textsc{TOFU}).}
    \begin{tabular}{llccccc}
        \toprule
        LLM & Method & Node $\downarrow$ & Edge $\downarrow$ & Node (others) $\uparrow$ & Edge (others) $\uparrow$ & General $\uparrow$ \\
        \midrule
        \multirow{5}{*}{Mistral-7B}
        & ICU & \textbf{0.0} & 18.9 & 0.0 & 18.9 & 58.8 \\
        & GA & \textbf{0.0} & \textbf{0.0} & 38.9 & 67.9 & 60.0\\ 
        & NPO & \textbf{0.0} & 5.4 & 38.9 & 61.3 & 62.4 \\
        & Ours (+GA) & \textbf{0.0} & \textbf{0.0} & \textbf{72.2} & 69.4 & 61.5 \\
        & Ours (+NPO) & \textbf{0.0} & \textbf{0.0} & 66.7 & \textbf{83.5} & \textbf{62.9} \\
        \midrule
        \multirow{5}{*}{Llama3.1-8B}
        & ICU & \textbf{0.0} & 67.4 & 0.0 & 67.4 & 59.5  \\
        & GA & \textbf{0.0} & \textbf{0.0} & 6.7 & 4.1 & 51.4 \\
        & NPO & \textbf{0.0} & 2.3 & 17.8 & 17.4 & 57.7 \\
        & Ours (+GA) & \textbf{0.0} & \textbf{0.0} & \textbf{55.6} & 61.1 & 60.1 \\
        & Ours (+NPO) & \textbf{0.0} & \textbf{0.0} & 48.9 & \textbf{70.2} & \textbf{60.2} \\
        \bottomrule
    \end{tabular}
    \label{tofu_results}
    \end{subtable}
\end{table*}

\subsection{Evaluation on Fine-Tuned Knowledge}
Next, we evaluate unlearning in a fine-tuning context, which reflects real-world scenarios where LLMs are customized with proprietary data. 
This setting is especially relevant for deleting sensitive information that may have been memorized during such fine-tuning.
We leverage the fictional \textsc{TOFU} dataset in this setting.
We fine-tune the LLM on the explanatory corpus of 20 \textsc{TOFU} entities, then apply unlearning for each target.
Table~\ref{tofu_results} shows that both variants of our method effectively remove the fine-tuned knowledge and outperform the baselines in preserving non-target knowledge and general utility. 
As a reference, the general utility scores of the pre-unlearned LLM are 62.6\% for Mistral and 60.4\% for Llama. 
These findings demonstrate that our method generalizes beyond pre-trained knowledge and remains effective even when the target knowledge originates entirely from fine-tuning.
This generalization highlights the practicality of our approach in post-deployment settings, where fine-tuned knowledge may need to be removed.

\subsection{Ablation: Effectiveness of Each Loss}
To better understand the effectiveness of the two losses in our method, we perform an ablation study focusing on them.
Specifically, we evaluate two variants: one that uses only the triplet-based loss ($\mathcal{L}_1$-only) and one that uses only the sentence-based loss ($\mathcal{L}_2$-only).
In this ablation, both losses leverage GA as the optimization algorithm.
In the $\mathcal{L}_1$-only setting, we continue unlearning until the unlearned LLM can no longer generate valid knowledge triplets about the forgetting target, at which point the process naturally stops.
Table~\ref{results_objective} shows that the $\mathcal{L}_1$-only variant is effective at suppressing attributional knowledge (EdgeAcc: 23.9\% in Mistral and 37.6\% in Llama) but less effective at removing the target entity itself (NodeAcc: 80.1\% in Mistral and 55.3\% in Llama).  
While the $\mathcal{L}_1$-only variant cannot achieve complete forgetting, we can confirm that the triplet-based loss contributes more to achieving EU.
Conversely, the $\mathcal{L}_2$-only variant achieves complete NU of the forgetting target (NodeAcc: 0.0\% for both LLMs) and is effective for EU (EdgeAcc: 0.0\% in Mistral and 4.0\% in Llama). 
However, it results in much lower non-target accuracy and general utility, suggesting that relying solely on sentence-based loss can lead to overly aggressive forgetting.  
This is because the sentence-based loss struggles to fully suppress attributional knowledge, requiring more updates, and increasing the damage to unrelated knowledge. 
Combining both losses (Full) achieves complete forgetting on NodeAcc and EdgeAcc while preserving the non-target knowledge and general utilities. \par
These results confirm that the two losses play complementary roles: the triplet-based loss accelerates the suppression of attributional knowledge, while the sentence-based loss promotes the forgetting of general information about the target entity.
By combining both losses, our full method leverages the strengths of each and achieves more balanced, effective, and selective unlearning.

\begin{table*}[tb]
\small
    \centering
    \caption{\textbf{Ablation results evaluating each loss.}
    The unlearning performance of using only the triplet-based loss ($\mathcal{L}_1$-only), only the sentence-based loss ($\mathcal{L}_2$-only), and their combination (Full).
Metrics include target forgetting (Node $\downarrow$, Edge $\downarrow$), non-target preservation (Node (others) $\uparrow$, Edge (others) $\uparrow$), and general utility (General $\uparrow$).}
    \begin{tabular}{llccccc}
        \toprule
        LLM & Method & Node $\downarrow$ & Edge $\downarrow$ & Node (others) $\uparrow$ & Edge (others) $\uparrow$ & General $\uparrow$ \\
        \midrule
        \multirow{3}{*}{Mistral-7B}
        & $\mathcal{L}_1$-only & 80.1 & 23.9 & 83.5 & 95.9 & 61.3 \\
        & $\mathcal{L}_2$-only & 0.0 & 0.0 & 17.0 & 33.1 & 51.0 \\
        & Full & 0.0 & 0.0 & 69.8 & 84.0 & 59.1 \\
        \midrule
        \multirow{3}{*}{Llama3.1-8B}
        & $\mathcal{L}_1$-only & 55.3 & 37.6 & 72.0 & 92.4 & 57.6 \\
        & $\mathcal{L}_2$-only & 0.0 & 4.0 & 30.2 & 54.8 & 55.5 \\
        & Full & 0.0 & 0.0 & 49.3 & 70.3 & 55.1 \\
        \bottomrule
    \end{tabular}
    \label{results_objective}
\end{table*}

\begin{table*}[tb]
\small
    \centering
    \caption{\textbf{Impact of KG Coverage on unlearning performance.}
    We compare two variants of our method: one using standard triplet extraction and one with relation-aware extraction (Rel-Aware).
    We report unlearning performance (Node $\downarrow$, Edge $\downarrow$), non-target preservation (Node (others) $\uparrow$, Edge (others) $\uparrow$), and general performance (General $\uparrow$).
    We also include KG Coverage, which measures the percentage of relevant attributions captured in the generated triplets during the process.}
    \begin{tabular}{llcccccc}
        \toprule
        LLM & Method & Node $\downarrow$ & Edge $\downarrow$ & Node (others) $\uparrow$ & Edge (others) $\uparrow$ & General $\uparrow$ & KG Coverage $\uparrow$ \\
        \midrule
        \multirow{2}{*}{Mistral-7B}
        & Default & 0.0 & 0.0 & 69.8 & 84.0 & 59.1 & 38.0 \\ 
        & Rel-Aware & 0.0 & 0.0 & 74.2 & 89.1 & 60.6 & 88.0 \\
        \midrule
        \multirow{2}{*}{Llama3.1-8B}
        & Default & 0.0 & 0.0 & 49.3 & 70.3 & 55.1 & 42.6 \\ 
        & Rel-Aware & 0.0 & 0.0 & 66.9 & 85.7 & 57.6 & 87.1 \\
        \bottomrule
    \end{tabular}
    \label{results_coverage}
\end{table*}

\subsection{Additional Study: Impact of KG Coverage}
\label{kt_coverage}
Next, we evaluate the KG coverage and its effect on our method.
Here, we define KG coverage as the proportion of the target entity's triplets successfully captured in the generated triplets during the process.
Our analysis considers only those triplets that the pre-unlearned LLM could correctly answer, ensuring we measure the knowledge that the LLM stores.
Specifically, we address two key questions:
(1) How many of the target's knowledge triplets are captured during the process?
(2) How does this KG coverage affect the effectiveness of our method?
To explore these questions, we compare our default \textsc{Get\_Attr} function with a relation-aware variant (Rel-Aware) explicitly specifying relation types in the \textsc{Get\_Attr} procedure.
Concretely, Rel-Aware constructs prompts such that each generation is conditioned not only on the target entity but also on a specific relation type, e.g., prompting the LLM to complete a triplet like (``Harry Potter'', ``defeat'', -) by generating the appropriate object.
This encourages the LLM to systematically enumerate facts associated with each relation, thereby improving KG coverage.
The list of relation types is taken from the \textsc{ParaRel} dataset, which provides a diverse set of predefined factual relations.
Further implementation of the Rel-Aware is provided in Appendix~\ref{get_attr_rel}. \par
Table~\ref{results_coverage} shows that while both settings result in complete unlearning (NodeAcc: 0.0\%, EdgeAcc: 0.0\%), the Rel-Aware variant achieves substantially higher KG coverage, nearly 90\%.
This confirms that providing relevant relation types in the \textsc{Get\_Attr} function dramatically improves the recall of comprehensive facts.
Importantly, this improved coverage leads to more selective unlearning, improving the preservation of non-target knowledge and general utility.
These findings highlight the importance of comprehensive triplet extraction in our method.
However, predefining all the relation types may not be feasible in real scenarios.
This trade-off between coverage and practicality motivates the future work on automated extraction that can balance completeness with practical constraints.
We also provide an analysis of KG precision, which measures the correctness of the extracted triplets, along with concrete examples of triplets extracted during the process, in Appendix~\ref{kt_precision}.

\section{Limitations}
While our proposed method demonstrates strong performance in achieving CU, several limitations remain.
First, our experiments show that the method can cause some unintended damage to non-target knowledge. Mitigating such collateral damage while maintaining effective unlearning remains an important challenge for future work.
Second, our current CU formulation is restricted to one-hop knowledge triplets, which may not easily capture more complex forms of knowledge, such as procedural knowledge.
However, this choice follows a widely adopted formalization of LLM knowledge and probing~\citep{Petroni2019LanguageMA,jiang-etal-2020-know,meng-etal-2022-rewire}.
Moreover, unlike prior research that assumes access to explicit knowledge, our approach does not require target triplets to be provided. 
Instead, we self‑generate candidate triplets for the forgetting target directly from the model and use them as structured unlearning targets, better aligning with the model’s internal knowledge.

\section{Conclusion}
In this paper, we propose a new requirement for LLM unlearning, called Concept Unlearning (CU), which targets concept-level knowledge rather than individual training data points.
To achieve CU, we introduced a novel self-supervised method in which the LLM generates its knowledge representations, both as triplets and explanatory sentences, and then applies the unlearning process to these self-constructed representations.
Through experiments on both real-world and synthetic datasets, we demonstrated that our approach consistently outperforms existing baselines regarding forgetting performance and selectivity.
We further analyzed how each component of our method contributes to forgetting performance and explored the factors influencing its effectiveness.
While our approach assumes triplet-based knowledge representations, it opens avenues for future work on unlearning more complex, implicit, or distributed forms of knowledge.
We hope this work provides a foundation for more practical LLM unlearning frameworks and contributes to realizing more reliable and trustworthy LLM systems.

\bibliography{aaai2026} 

\appendix
\section{Full Results}
\subsection{Evaluation on Other LLMs}
\label{other_llms}
To assess the generality of our approach, we additionally apply Ours (+GA) to two other LLMs: Qwen2.5-7B Instruct~\citep{yang2024qwen2} and Gemma2-9B Instruct~\citep{gemma_2024}.
We follow the same evaluation protocol as described in the main experiments and evaluate CU metrics (NodeAcc, EdgeAcc, and their non-target counterparts) and general utility.
In this experiment, we evaluate all methods on the \textsc{Wiki-Fact} corpus and compare Ours (+GA) against two representative baselines: ICU and GA on the \textsc{Wiki-Fact} corpus. \par
The results are shown in Table~\ref{appendix_results}.
Ours (+GA) consistently achieves complete unlearning for Gemma2.
While Ours (+GA) cannot achieve complete forgetting in Qwen2.5, it shows strong suppression for the forgetting target.
In addition, compared to GA on the \textsc{Wiki-Fact} dataset, our method achieves substantially higher non-target preservation and general utility, confirming the selective forgetting property.
For example, in the case of Qwen2.5, GA achieves complete unlearning but suffers significant collateral damage (i.e., NodeAcc (others): 11.5\% and EdgeAcc (others)= 11.1\%), whereas Ours (+GA) maintains much higher non-target accuracy (i.e., NodeAcc (others)=48.7\% and EdgeAcc (others)= 67.6\%).
A similar trend is observed for Gemma2, where Ours (+GA) outperforms ICU and GA in preserving utility while achieving complete forgetting.
These results align well with the main experiments on Mistral and Llama3.1, demonstrating that our method generalizes effectively across LLM architectures. \par
\begin{figure*}[tb]
\begin{minipage}{0.32\textwidth}
\begin{center}
\includegraphics[width=\textwidth]{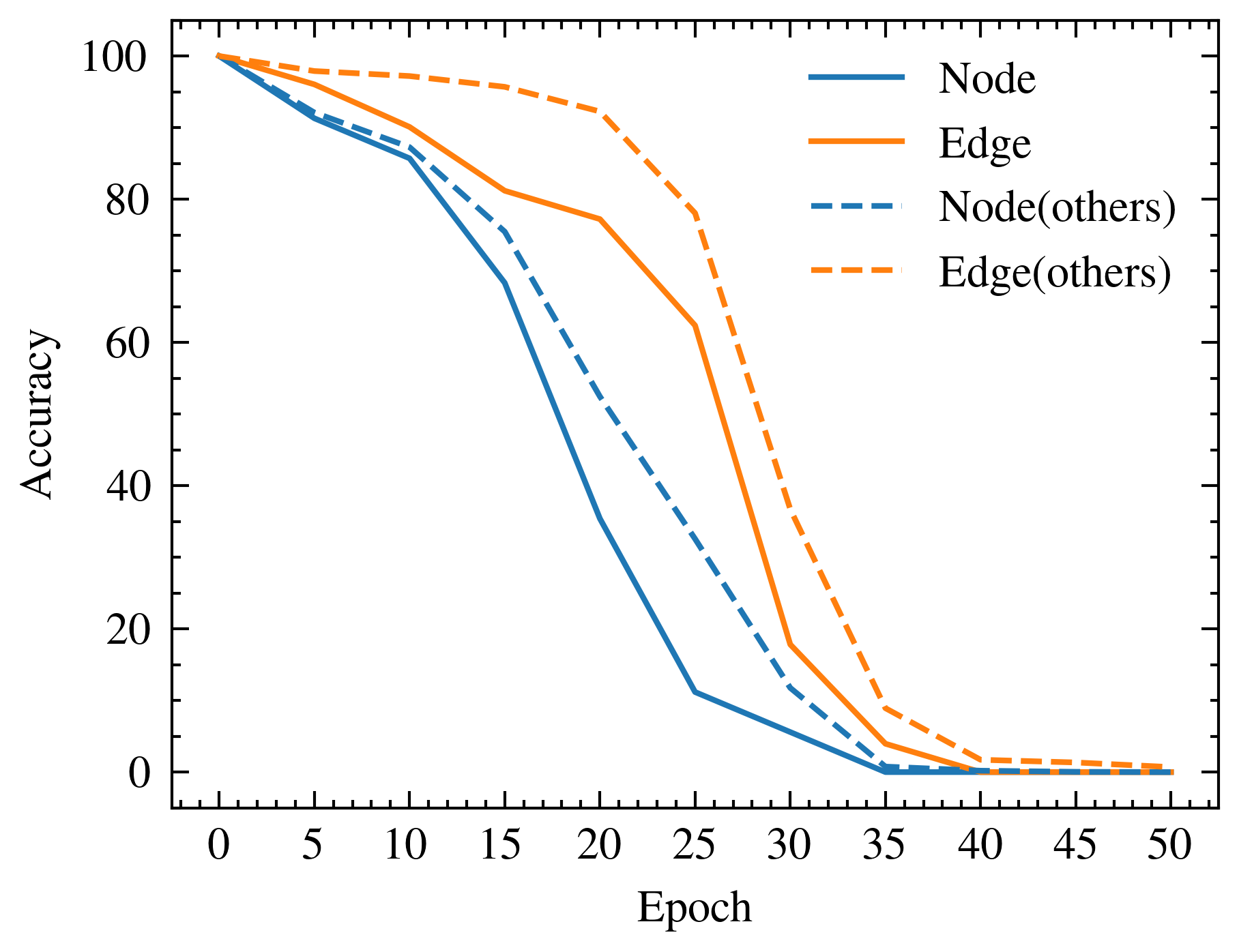}
\end{center}
\subcaption{Results on Llama3.1-8B}
\end{minipage}
\begin{minipage}{0.32\textwidth}
\begin{center}
\includegraphics[width=\textwidth]{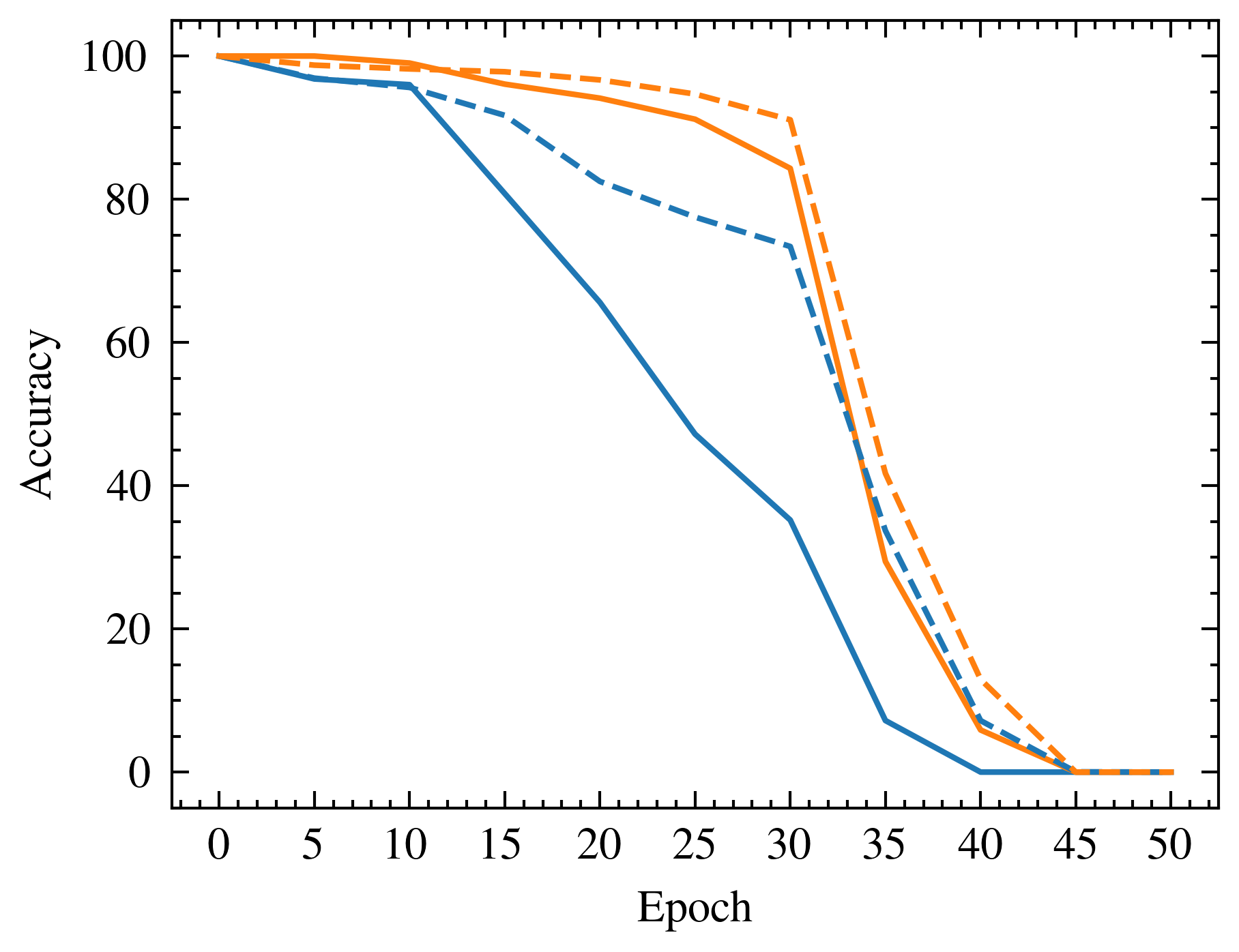}
\end{center}
\subcaption{Results on Qwen2.5-7B}
\end{minipage}
\begin{minipage}{0.32\textwidth}
\begin{center}
\includegraphics[width=\textwidth]{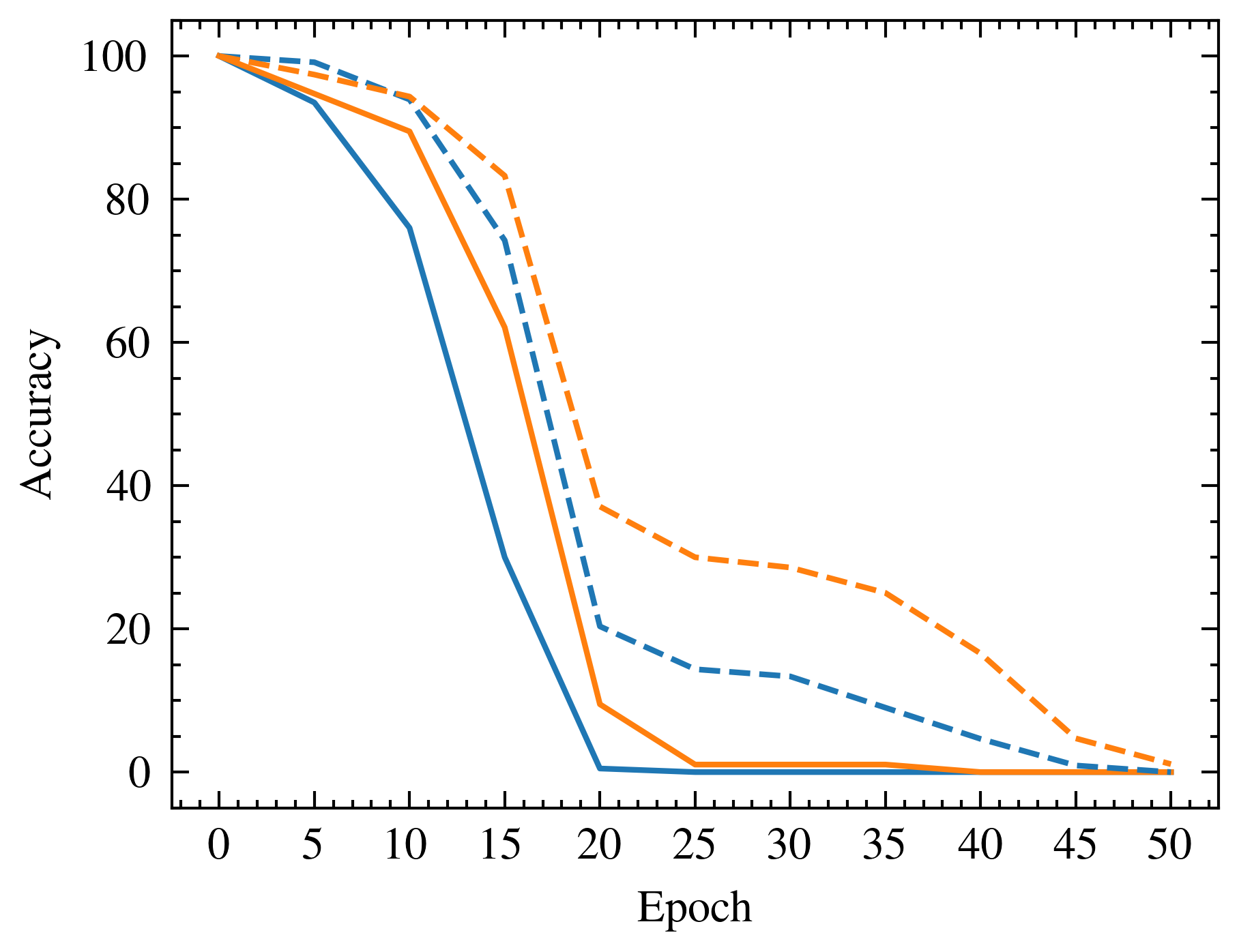}
\end{center}
\subcaption{Results on Gemma2-9B}
\end{minipage}
\caption{\textbf{Results of GA on the \textsc{Wiki-Fact} corpus.}
The blue and orange lines represent NodeAcc and EdgeAcc, respectively.  
Solid lines indicate performance on target entities, while dashed lines correspond to non-target entities.  
Each plot shows the forgetting dynamics under naive GA for a different model: Llama3.1-8B (left), Qwen2.5-7B (center), and Gemma2-9B (right). 
}
\label{naive-GA-other}
\end{figure*}
\begin{figure*}[tb]
\begin{minipage}{0.32\textwidth}
\begin{center}
\includegraphics[width=\textwidth]{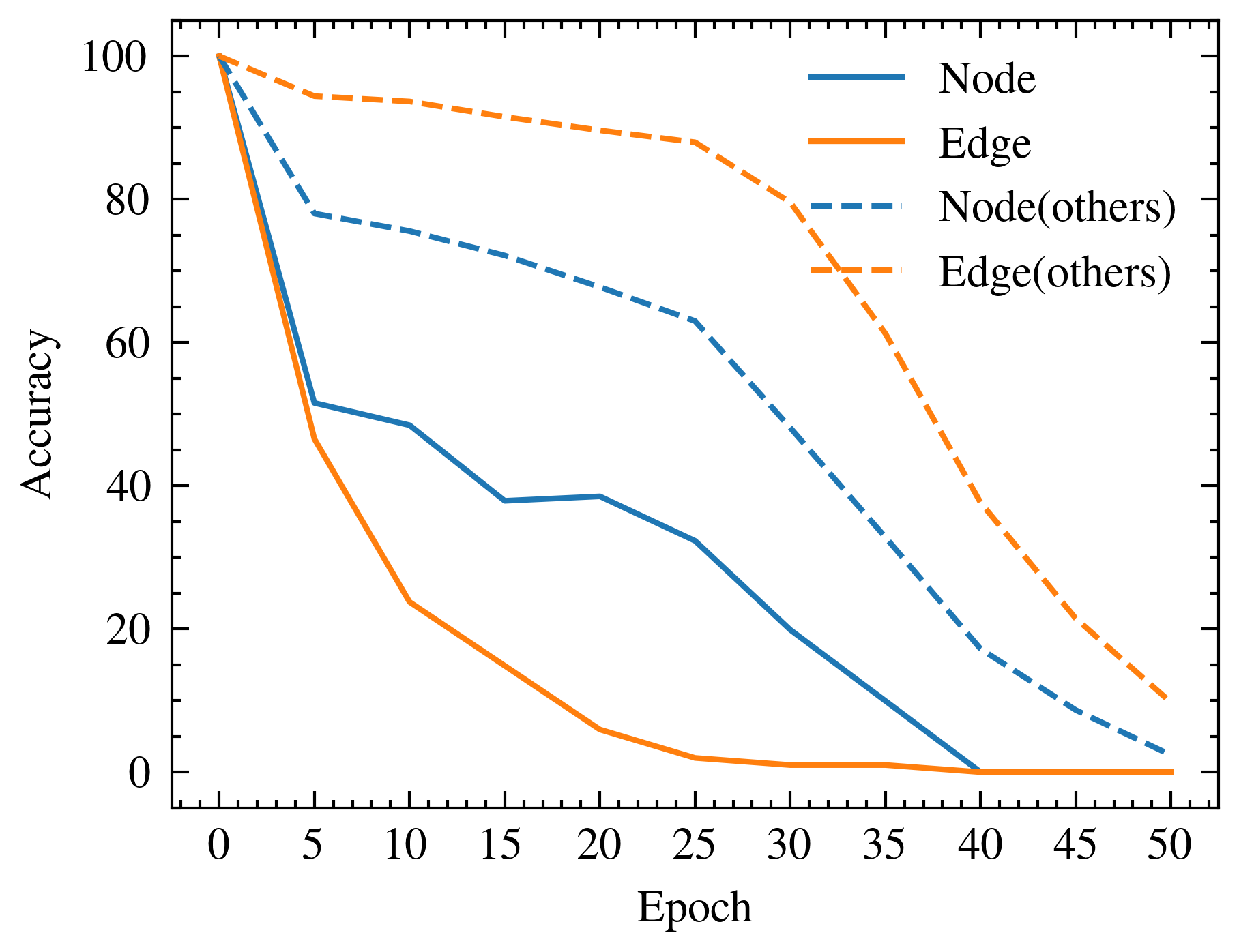}
\end{center}
\subcaption{Results on Llama3.1-8B}
\end{minipage}
\begin{minipage}{0.32\textwidth}
\begin{center}
\includegraphics[width=\textwidth]{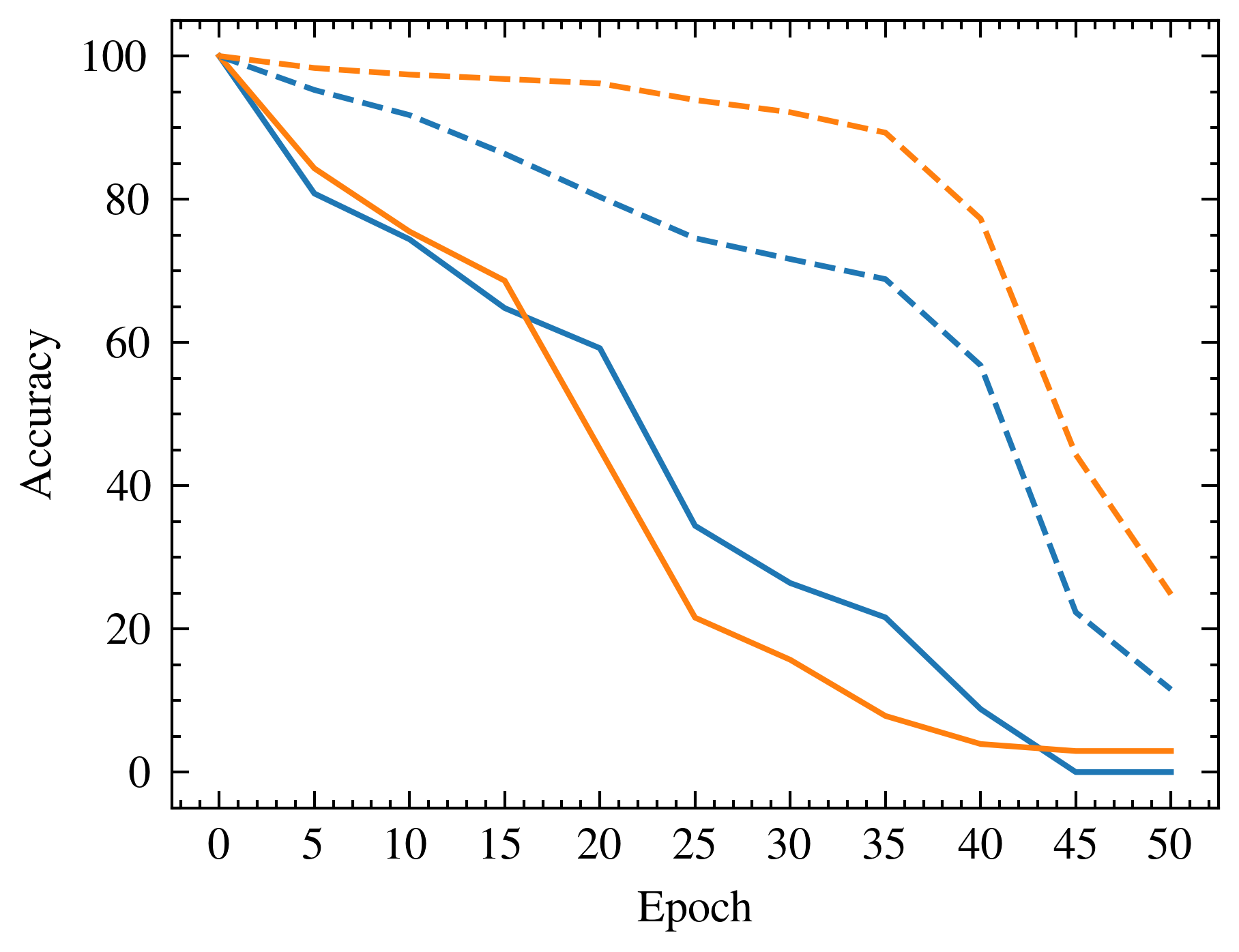}
\end{center}
\subcaption{Results on Qwen2.5-7B}
\end{minipage}
\begin{minipage}{0.32\textwidth}
\begin{center}
\includegraphics[width=\textwidth]{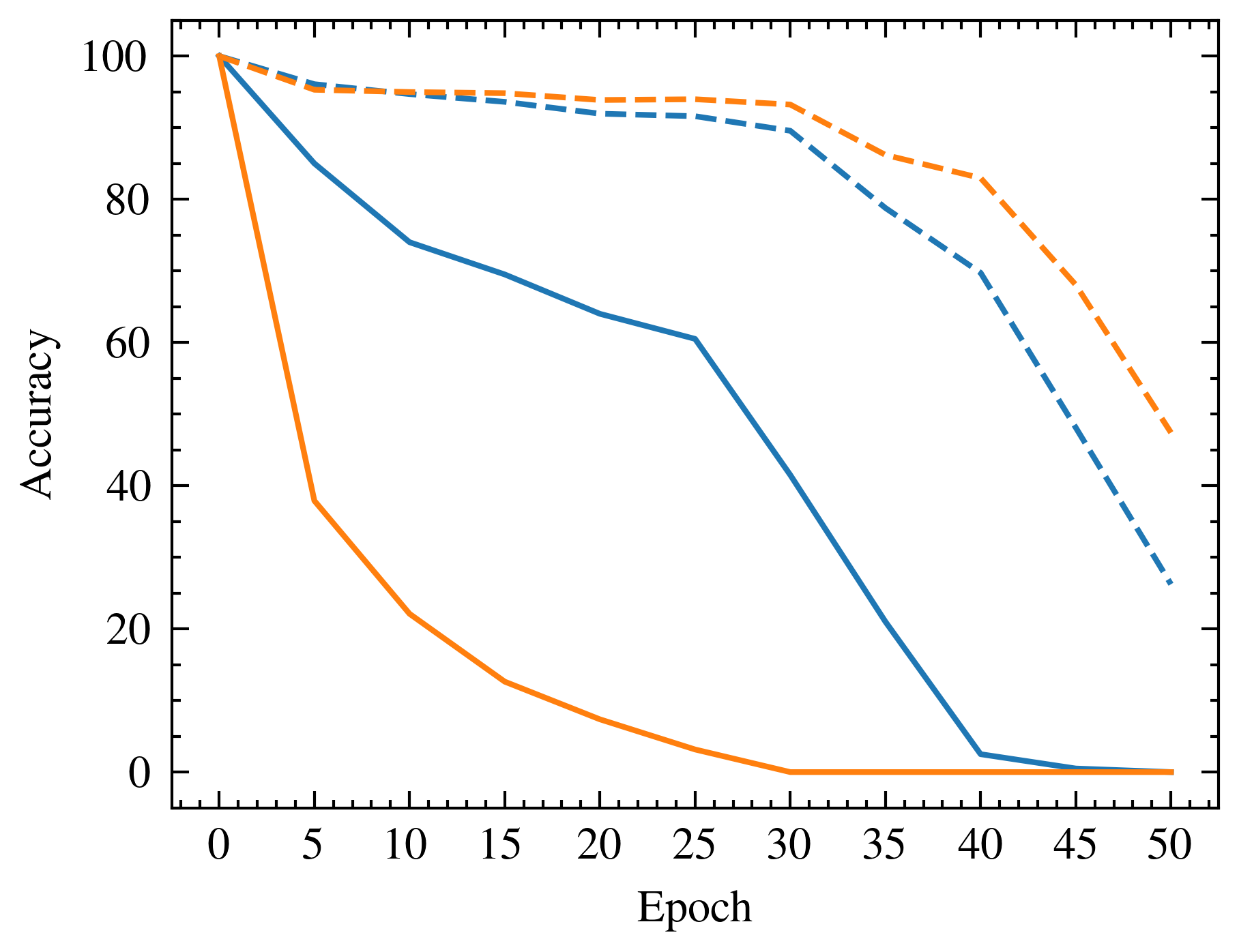}
\end{center}
\subcaption{Our method on Gemma2-9B}
\end{minipage}
\caption{\textbf{Results of Ours (+GA) on \textsc{Wiki-Fact} knowledge.}
The blue and orange lines represent NodeAcc and EdgeAcc, respectively.  
Solid lines indicate performance on target entities, while dashed lines correspond to non-target entities.  
Each plot shows the forgetting dynamics under our method for a different model: Llama3.1-8B (left), Qwen2.5-7B (center), and Gemma2-9B (right).
}
\label{cu_evolution_ours_other}
\end{figure*}

We also show the evolution of CU metrics during the GA-based unlearning process on Llama3.1, Qwen2.5, and Gemma2.
Figure~\ref{naive-GA-other} shows the results of naive GA applied to the \textsc{Wiki-Fact} corpus.
It shows a similar trend to Fig.~\ref{CU_evolution_GA} that EdgeAcc for the forgetting targets decays only gradually, and accuracy for both targets and non-targets declines simultaneously as unlearning progresses.
Therefore, we can confirm that the naive GA on the \textsc{Wiki-Fact} corpus lacks selectivity.
On the other hand, Fig.~\ref{cu_evolution_ours_other} demonstrates that Ours (+GA) reduces EdgeAcc for the forgetting targets rapidly, and the accuracy degradation for non-targets is modest, suggesting that our method achieves effective CU on different LLM architectures, as with the main results in Fig.~\ref{CU_evolution_ours}.

\begin{table*}[tb]
\footnotesize
    \centering
    \caption{\textbf{Results on \textsc{Wiki-Fact} dataset.}
    The unlearning performance on Qwen2.5-7B and Gemma2-9B.  
Metrics include forgetting target accuracy (Node $\downarrow$, Edge $\downarrow$), non-target knowledge preservation (Node (others) $\uparrow$, Edge (others) $\uparrow$), and general utility (General $\uparrow$).  
The best value in each column is shown in bold.
    }
    \begin{tabular}{llccccc}
        \toprule
        LLM & Method & Node $\downarrow$ & Edge $\downarrow$ & Node (others) $\uparrow$ & Edge (others) $\uparrow$ & General $\uparrow$ \\
        \midrule
        \multirow{3}{*}{Qwen2.5-7B}
        & ICU & 48.8 & 93.1 & \textbf{53.7} & \textbf{89.5} & \textbf{59.6} \\
        & GA & \textbf{0.0} & \textbf{0.0} & 11.5 & 11.1 & 41.2 \\
        & Ours (+GA) & \textbf{0.0} & 2.94 & 48.7 & 67.6 & 56.9 \\
        \midrule
        \multirow{3}{*}{Gemma2-9B}
        & ICU & 39.8 & 95.8 & 30.4 & \textbf{90.9} & \textbf{63.8}  \\
        & GA & \textbf{0.0} & \textbf{0.0} & 25.1 & 38.3 & 45.1 \\
        & Ours (+GA) & \textbf{0.0} & \textbf{0.0} & \textbf{76.3} & 83.8 & 59.6  \\
        \bottomrule
    \end{tabular}
    \label{appendix_results}
\end{table*}

\subsection{Task-wise Results for General Utility Evaluation}
\label{gen_results}
To complement the average general utility scores reported in the main results, we provide task-wise breakdowns across seven standard benchmarks: HellaSwag, Lambada, Winogrande, COPA, ARCEasy, ARCChallenge, and PubmedQA~\citep{zellers-etal-2019-hellaswag,kazemi-etal-2023-lambada,sakaguchi2021winogrande,copa,Clark2018ThinkYH,jin-etal-2019-pubmedqa}.
HellaSwag and Lambada are benchmarks of linguistic reasoning ability, Winogrande and COPA are benchmarks of commonsense-based reasoning ability, and ARC-Easy, ARC-Challenge, and PubmedQA are benchmarks of scientific reasoning ability.
These results offer a detailed view of how each unlearning method affects specific reasoning capabilities of the LLMs.\par
\begin{figure*}[tb]
\begin{minipage}{0.48\textwidth}
\begin{center}
\includegraphics[width=\textwidth]{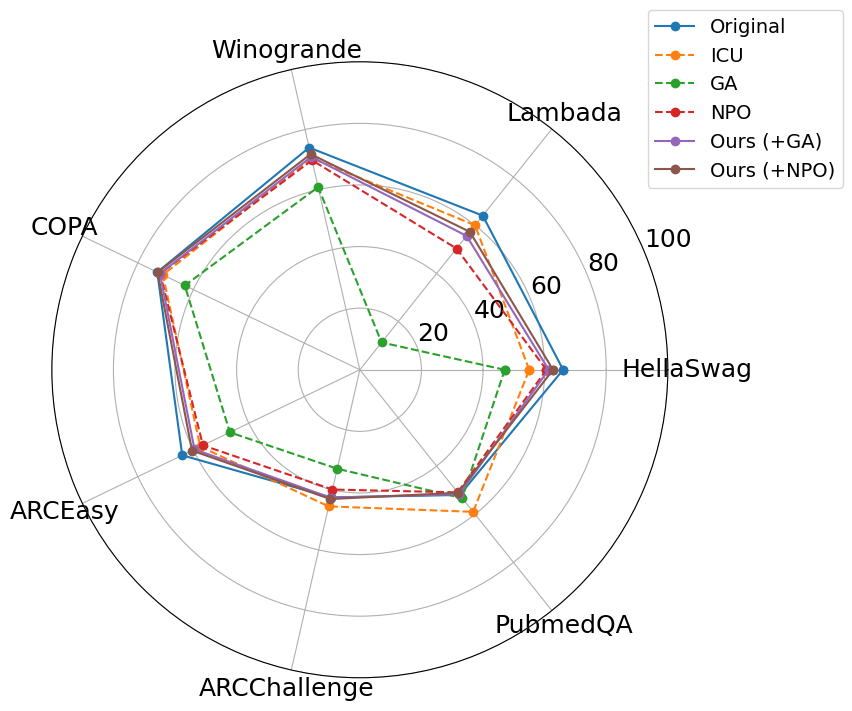}
\end{center}
\subcaption{Results on Mistral}
\end{minipage}
\begin{minipage}{0.48\textwidth}
\begin{center}
\includegraphics[width=\textwidth]{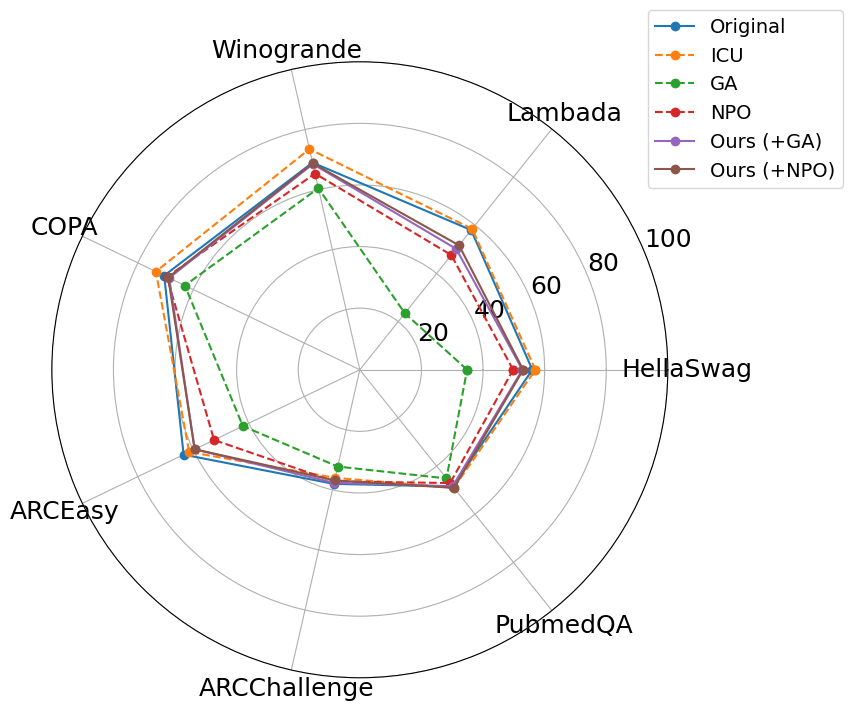}
\end{center}
\subcaption{Results on Llama3.1}
\end{minipage}
\caption{\textbf{General Utility Across Tasks After Unlearning \textsc{Wiki-Fact} Knowledge}.
Radar plots show performance across seven inference tasks for each method.
Solid lines represent the original LLM and our proposed methods (Ours (+GA), Ours (+NPO)), while dashed lines represent the baselines (ICU, GA, NPO).
Our methods perform comparably to the original LLM, indicating minimal degradation in general reasoning capabilities.}
\label{general_utility_wiki}
\end{figure*}
\begin{figure*}[tb]
\begin{minipage}{0.48\textwidth}
\begin{center}
\includegraphics[width=\textwidth]{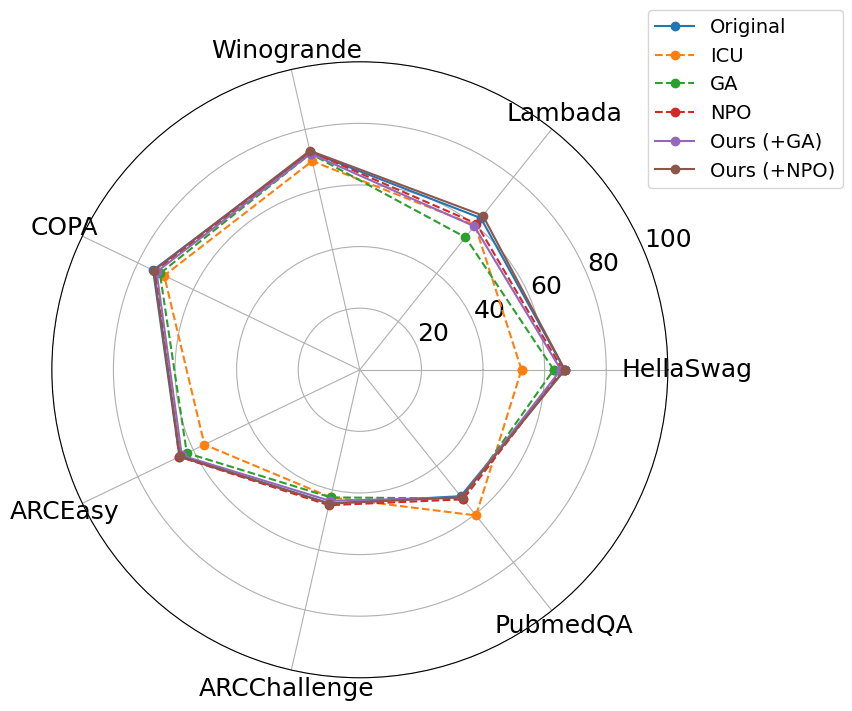}
\end{center}
\subcaption{Results on Mistral}
\end{minipage}
\begin{minipage}{0.48\textwidth}
\begin{center}
\includegraphics[width=\textwidth]{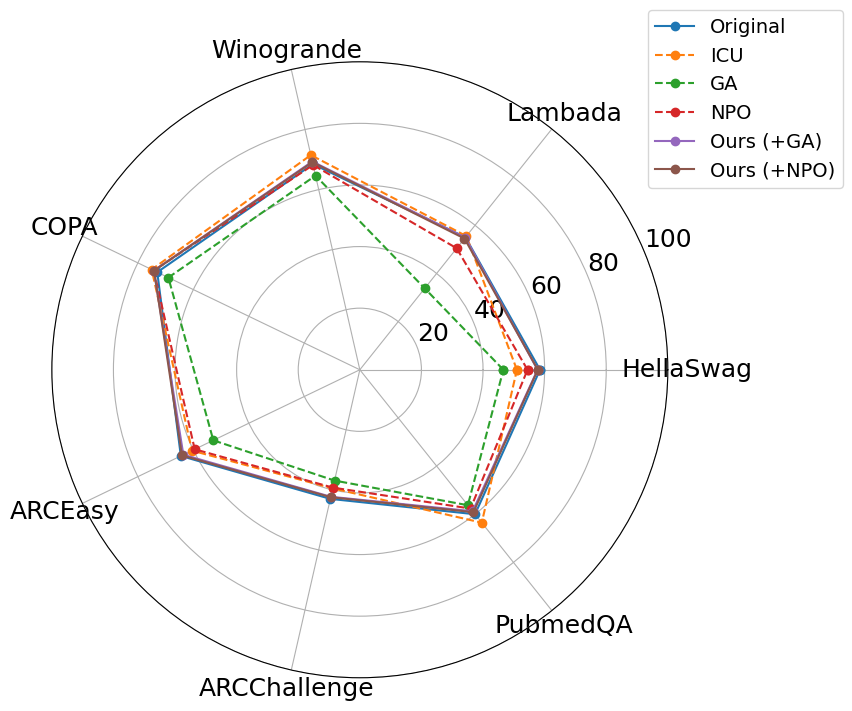}
\end{center}
\subcaption{Results on Llama3.1}
\end{minipage}
\caption{\textbf{General Utility Across Tasks After Unlearning \textsc{TOFU} Knowledge}.  
Radar plots show performance across seven inference tasks for each method.  
Solid lines represent the original LLM and our proposed methods (Ours (+GA), Ours (+NPO)), while dashed lines represent the baselines (ICU, GA, NPO).  
Our methods maintain strong general performance while effectively unlearning fine-tuned knowledge.}
\label{general_utility_tofu}
\end{figure*}

Figures~\ref{general_utility_wiki} and~\ref{general_utility_tofu} present radar plots for Mistral and Llama3.1 across all seven benchmarks, evaluated after unlearning \textsc{Wiki-Fact} and \textsc{TOFU} knowledge, respectively.
We compare the original LLMs with ICU, GA, NPO, and our two variants: Ours (+GA) and Ours (+NPO).
As shown, our methods maintain general utility close to the pre-unlearned LLMs across tasks, demonstrating that our approach avoids collateral damage to unrelated reasoning capabilities.
This aligns with the high average utility scores reported in the main results.
Both NPO and GA on the external corpus show strong forgetting performance on the target concepts, effectively suppressing memorized knowledge. 
However, these baselines also lead to a larger degradation in general utility metrics compared to our method. 
This indicates that while they succeed in forgetting, they tend to compromise the LLM's broader reasoning ability, highlighting the superior balance our approach achieves between targeted forgetting and the preservation of general capabilities.
On the other hand, ICU maintains general utility nearly at the level of the pre-unlearned LLMs, but its forgetting performance is limited, especially in suppressing attribute-level knowledge, highlighting a trade-off between the precision of forgetting and the retention of general capabilities.
These comparisons underscore the strength of our method in balancing effective concept unlearning with minimal loss in reasoning ability. \par
\begin{figure*}[tb]
\begin{minipage}{0.48\textwidth}
\begin{center}
\includegraphics[width=\textwidth]{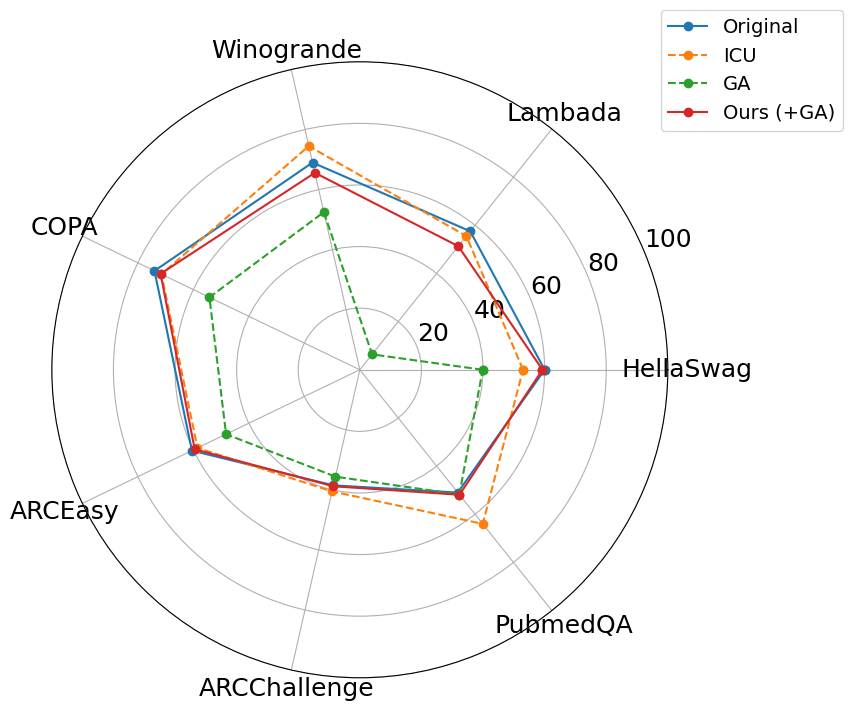}
\end{center}
\subcaption{Results on Qwen2.5}
\end{minipage}
\begin{minipage}{0.48\textwidth}
\begin{center}
\includegraphics[width=\textwidth]{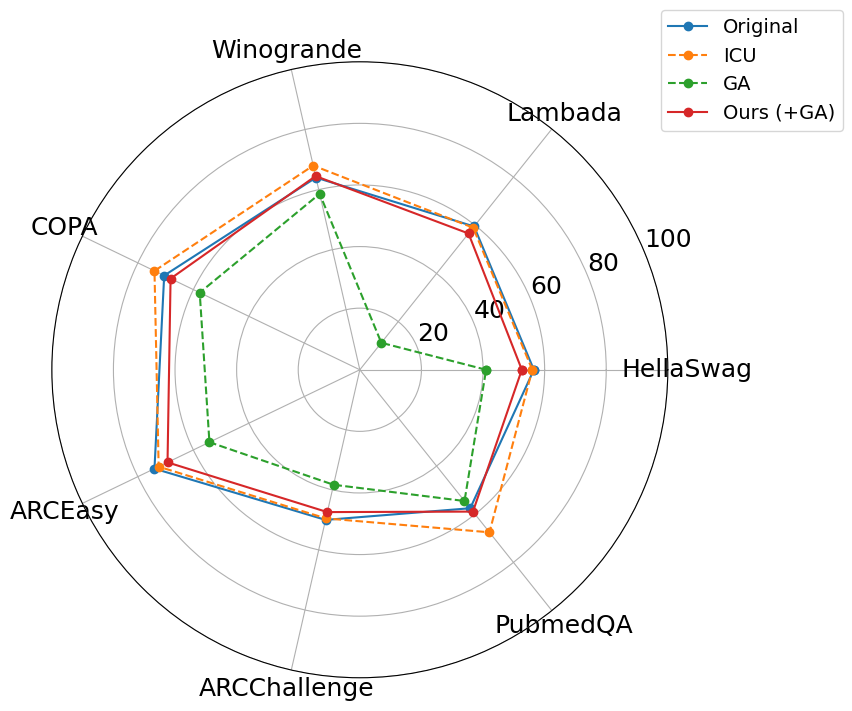}
\end{center}
\subcaption{Results on Gemma2}
\end{minipage}
\caption{\textbf{General Utility Across Tasks After Unlearning \textsc{Wiki-Fact} Knowledge}.
Radar plots show performance across seven inference tasks for each method.
Solid lines represent the original LLM and our proposed methods (Ours (+GA)), while dashed lines represent the baselines (ICU, GA).
Our methods perform comparably to the original LLM, indicating minimal degradation in general reasoning capabilities.}
\label{general_utility_wiki_other}
\end{figure*}

Additionally, Fig.~\ref{general_utility_wiki_other} reports task-wise general utility results for two other LLMs: Qwen2.5-7B and Gemma2-9B, under the \textsc{Wiki-Fact} dataset.
These evaluations compare the pre-unlearned LLM, ICU, GA, and Ours (+GA).
As in the previous results, GA significantly degrades general utility despite strong forgetting performance, while ICU preserves utility but fails to fully forget the target knowledge.
In contrast, Ours (+GA) consistently achieves complete forgetting with minimal utility loss, further demonstrating its robustness across LLM architectures.
These results further support the robustness and general applicability of our method across different LLM architectures.

\subsection{Additional Study: Impact of KG Precision in Our Method}
\label{kt_precision}
Besides the KG coverage studied in Sec.~\ref{kt_coverage}, we hypothesize that the precision of the knowledge graph, i.e., the fraction of generated triplets that are factually correct, may also play a critical role in effective CU.
In this section, we quantify the KG precision and examine its potential influence on unlearning performance.
Here, we define KG precision as the proportion of factually correct triplets among all triplets generated during the unlearning process.
This metric captures how accurate the generated knowledge is with respect to real-world facts and reflects the reliability of the extracted triplets.\par
To measure KG precision, we leverage GPT-4 as an external verifier.
For each triplet $(\bm{s}, \bm{r}, \bm{o})$, we prompt GPT-4: ``Tell me if the knowledge triplet is correct. If correct, say 'yes'; if not, say 'no'. $(\bm{s}, \bm{r}, \bm{o})$.''.
A ``yes'' labels the triplet as valid.
KG precision is the percentage of such valid triplets among all triplets generated during the unlearning process.
Through this evaluation, we address two key questions: (1) How accurate are the triplets generated during unlearning?
(2) How does that accuracy affect the effectiveness of our method?
To explore these questions, we compare our default pipeline, where triplet validation in the \textsc{Get\_Attr} function is conducted using the pre-unlearned LLM, with a variant that replaces this validation step with GPT-4o (denoted as w/ GPT-4o).
Given that GPT-4o consistently outperforms mid-sized LLMs such as Llama3.1-8B or Mistral-7B on factual QA benchmarks, incorporating GPT-4o for triplet validation is expected to improve the KG precision. \par
The results in Table~\ref{results_precision} show that both settings achieve perfect unlearning (NodeAcc: 0.0\%, EdgeAcc: 0.0\%).
The w/ GPT-4o variant significantly improves the KG precision (i.e., 55-69\% to about 96\%) for both LLMs, confirming that a stronger verifier is highly effective at filtering erroneous triplets.
However, higher KG precision does not necessarily contribute to better selective unlearning in our method.
For Llama3.1-8B, using the cleaner triplet set significantly improves non-target accuracy and slightly enhances general utility, indicating that the larger LLM benefits from removing noisy knowledge during the unlearning process.
In contrast, for Mistral-7B, whose default KG precision was already higher, the same filtering yields virtually no gain, and even a small drop, in non-target scores.
Taken together with the coverage additional study in Sec.~\ref{kt_coverage}, these findings suggest that a nuanced trade-off: while maintaining sufficient KG precision is important, once a reasonable threshold is met, expanding KG coverage becomes the more influential factor for effective CU.

\begin{table*}[tb]
    \centering
    \caption{\textbf{Impact of KG Precision on unlearning performance.}
    We compare two variants of our method: one using the pre-unlearned LLM in \textsc{Get\_Attr} and one with GPT-4o in \textsc{Get\_Attr} (w/ GPT-4o).
    We report unlearning performance (Node $\downarrow$, Edge $\downarrow$), non-target preservation (Node (others) $\uparrow$, Edge (others) $\uparrow$), and general performance (General $\uparrow$).
    We also include KG Precision, which measures the percentage of correct triplets among all triplets generated during the unlearning process.
    }
    \scalebox{0.82}{
    \begin{tabular}{llcccccc}
        \toprule
        LLM & Method & Node $\downarrow$ & Edge $\downarrow$ & Node (others) $\uparrow$ & Edge (others) $\uparrow$ & General $\uparrow$ & KG Precision $\uparrow$ \\
        \midrule
        \multirow{2}{*}{Mistral-7B}
        & Default & 0.0 & 0.0 & 69.8 & 84.0 & 59.1 & 68.6 \\
        & w/ GPT-4o & 0.0 & 0.0 & 63.8 & 78.9 & 59.8 & 96.2 \\
        \midrule
        \multirow{2}{*}{Llama3.1-8B}
        & Default & 0.0 & 0.0 & 49.3 & 70.3 & 55.1 & 54.9 \\
        & w/ GPT-4o & 0.0 & 0.0 & 61.8 & 83.1 & 56.7 & 96.7  \\
        \bottomrule
    \end{tabular}
    \label{results_precision}
    }
\end{table*}

\subsection{Case Study: Evaluating Knowledge Triplets for \textit{Jesus Christ}}
\label{jesus_kt_analysis}
To better understand the quality and interpretability of the knowledge triplets captured during the unlearning process, we conduct a focused case study on the entity \textit{Jesus Christ} in the \textsc{Wiki-Fact} dataset.
We collect knowledge triplets generated by the LLM during the unlearning process using our proposed method (Ours (+GA)).
To evaluate the factual correctness and semantic centrality of each triplet $(\bm{e}^\mathrm{t}, \bm{r}, \bm{o})$, we use GPT-4 as an external annotator.
Specifically, we follow a two-step procedure:
\begin{enumerate}
    \item First, we prompt GPT-4 with: ``Tell me if the following knowledge triplet is correct. If correct, say yes; if not, say no: $(\bm{e}^\mathrm{t}, \bm{r}, \bm{o})$''. 
    If GPT-4 replies ``no'', the triplet is marked with a centrality score of $-1$ and excluded from further evaluation.
    \item If GPT-4 deems the triplet correct, we then prompt: ``Evaluate whether the following knowledge triple is central to the concept of Jesus Christ. Rate the centrality on a scale from 1 to 5, where: 5 = extremely central (core theological or biographical knowledge), 1 = peripheral (cultural, artistic, or indirect knowledge). Provide your score only. Knowledge Triplet: $(\bm{e}^\mathrm{t}, \bm{r}, \bm{o})$''. 
    The numerical score returned by GPT-4 is used as the centrality score.
\end{enumerate}
Note that the triplets are listed in the order in which they were generated by the LLM during the unlearning process.
If a triplet is generated multiple times, only the first occurrence is recorded and evaluated to avoid duplication. \par
Table~\ref{jesus_kt_eval} presents the complete set of knowledge triplets generated during the unlearning process, along with their corresponding centrality scores.
Out of 37 triplets generated, 35 are judged as factually correct by GPT-4.
A large proportion of these receive centrality scores of 4 or 5, indicating that the triplet extraction process tends to surface central pieces of knowledge stored in the LLM.
This supports the effectiveness of using self-generated triplets as a proxy for internal knowledge representations in the context of CU. \par
Interestingly, we observe a progressive decline in the scores as the unlearning process proceeds: while early triplets tend to score 4 or 5, later triplets more frequently receive lower scores of 2 or 3.
This trend suggests that the proposed method unlearns knowledge in a core to periphery fashion, first targeting highly central and salient facts, then gradually moving toward  more peripheral or marginal information.
Such a pattern implies that monitoring the centrality of extracted triplets could serve as an implicit signal of forgetting progress.
In particular, the point at which triplets shift predominantly to low-centrality may indicate a saturation point, where further unlearning might lead to diminishing returns or unnecessary damage to unrelated knowledge. 
This insight could inform an adaptive stopping criterion or feedback mechanism in future iterations of CU, potentially improving efficiency and selectivity in practical deployments.

\begin{table*}[tb]
\centering
\small
\caption{\textbf{Knowledge triplets generated for ``Jesus Christ'' during the unlearning process.} 
Each triplet was evaluated by GPT-4 for factual correctness and centrality to the concept of Jesus Christ.
Centrality scores range from 5 (high central, e.g., theological or biographical knowledge) to 1 (peripheral content), while -1 indicates factually incorrect or hallucinated information.
Triplets are listed in the order they were generated, with duplicates removed.}
\scalebox{0.9}{
\begin{tabular}{lllc}
\toprule
Subject & Relation & Object & Score (-1 to 5) \\
\midrule
Jesus Christ & religion or worldview & Christianity & 5 \\
Jesus Christ & birthplace & Bethlehem & 5 \\
Jesus Christ & teachings & Sermon on the Mount & 5 \\
Jesus Christ & crucifixion & Calvary & 5  \\
Jesus Christ & resurrection & Easter & 5  \\
Jesus Christ & disciples & twlve apostles & 5  \\
Jesus Christ & mother & Mary & 5 \\
Jesus Christ & father & Joseph & -1 \\
Jesus Christ & birth year & approximately 4 BC & 3 \\
Jesus Christ & death year & approximately 30 AD & 5 \\
Jesus Christ & miracles & turning water into wine, healing the blind, walking on water & 5 \\
Jesus Christ & significance & central figure of Christianity & 5 \\
Jesus Christ & teachings & love thy neighbor, forgive others & 5 \\
Jesus Christ & teachings & the Lord's Prayer & 5 \\
Jesus Christ & teachings & the Last Supper & -1 \\
Jesus Christ & miracles & healing the sick & 5  \\
Jesus Christ & miracles & raising the dead & 5 \\
Jesus Christ & miracles & multiplying loaves and fishes & 4 \\
Jesus Christ & miracles & calming the storm & 4 \\
Jesus Christ & miracles & casting out demons & 4 \\
Jesus Christ & miracles & healing the lepers & 4 \\
Jesus Christ & miracles & healing the paralyzed & 4 \\
Jesus Christ & miracles & healing the deaf & 4 \\
Jesus Christ & historical figure & Yes & 4 \\
Jesus Christ & resurrection & Yes & 5 \\
Jesus Christ & era & 1st century AD & 4 \\
Jesus Christ & biblical references & New Testament & 5 \\
Jesus Christ & symbols & Fish, Cross & 3 \\
Jesus Christ & holidays & Christmas, Easter & 4 \\
Jesus Christ & art & Depicted in various forms throughout history & 2 \\
Jesus Christ & impact & Influenced Western civilization & 3 \\
Jesus Christ & translations & Bible has been translated into over 600 languages & 2 \\
Jesus Christ & interpretations & Various theological and philosophical interpretations & 3 \\
\bottomrule
\end{tabular}
\label{jesus_kt_eval}
}
\end{table*}

\subsection{Why are Attributions Harder to Forget?}
Our main experiments in Sec.~\ref{evaluation} indicate that EU is generally more challenging than NU.
We hypothesize that this is due, in part, to the distributed nature of attributional knowledge across diverse relation types.
In this section, we explore an additional hypothesis: that the linguistic category of the token, specifically whether it is a proper or common noun, also affects its susceptibility to forgetting.
Since target entities are usually expressed as proper nouns, whereas attributional information is typically converted through common nouns, differences in word type may partially explain the observed gap in unlearning difficulty between NU and EU.
To test this hypothesis, we compare the change in token-level prediction loss for proper nouns and common nouns after applying a single GA update.
A larger increase in loss indicates that the token was more affected by the unlearning step (i.e., easier to forget), while a smaller increase implies greater resistance to forgetting.
Concretely, for each sentence in the \textsc{Wiki-Fact} dataset, we perform one GA update and then mask individual tokens one at a time.
We prompt the LLM to predict the masked token and measure the change in its prediction loss before and after the GA update.
To reduce confounding effects:
\begin{itemize}
  \item We restrict our analysis to tokens that appear only once in the sentence to avoid repetition effects.
  \item We consider only tokens corresponding to attribute values, not subject entities, ensuring functional and syntactic consistency.
\end{itemize}
We then classify the tokens into proper and common nouns and compare their average loss increases using Welch's t-test. 
As shown in Table~\ref{noun_results}, proper nouns consistently exhibit significantly larger increases in loss compared to common nouns across both Mistral-7B and Llama3.1-8B.
For Mistral-7B, the mean loss increase for proper nouns is $3.7 \times 10^{-2}$, more than double that for common nouns ($1.6 \times 10^{-2}$), with a statistically significant difference ($p = 0.042$). 
Similarly, for Llama3.1-8B, proper nouns again show higher sensitivity to forgetting with a mean loss delta of $9.3 \times 10^{-3}$, compared to $4.0 \times 10^{-3}$ for common nouns ($p = 0.015$).
This indicates that proper nouns, typically associated with entities, are more sensitive to GA-based unlearning, while common nouns, often carrying attributional information, are more resistant.
These findings support our hypothesis and suggest that, in addition to structural factors like distributedness, the linguistic nature of tokens also plays a role in unlearning difficulty.
Understanding these factors may help guide the development of more fine-grained and adaptive unlearning methods in future work.

\begin{table*}[tb]
\small
    \centering
    \caption{Results on \textsc{TOFU} dataset.}
    \begin{tabular}{lcccc}
        \toprule
        LLM & Noun Type & Mean (Std) & t-value & p-value  \\
        \midrule
        \multirow{2}{*}{Mistral 7B}
        & Proper &  $3.7\times 10^{-2}$ ($8.5\times 10^{-2}$) & \multirow{2}{*}{1.99} & \multirow{2}{*}{0.042} \\
        & Common & $1.6\times 10^{-2}$ ($4.3\times10^{-2}$) \\ 
        \midrule
        \multirow{2}{*}{Llama3.1 8B}
        & Proper & $9.3\times 10^{-3}$ ($1.7\times 10^{-2}$) & \multirow{2}{*}{2.40} & \multirow{2}{*}{0.015} \\
        & Common & $4.0\times 10^{-3}$ ($1.1\times 10^{-2}$) \\ 
        \bottomrule
    \end{tabular}
    \label{noun_results}
\end{table*}

\section{Experimental Details}
\label{eval_details}
\subsection{Experimental Setup}
For the main evaluation on the \textsc{Wiki-Fact} dataset, we conduct the forgetting process for each of the 30 target entities and measure the CU and the general utility metrics.  
To assess the preservation of non-target knowledge, we additinoally evaluate the CU metrics on 20 randomly selected entities not included in the forgetting targets.
Similarly, for the \textsc{TOFU} dataset, we conducted the forgetting process on 20 target entities and evaluated CU metrics.  
Non-target performance is measured using 19 distinct entities that are not selected as forgetting targets.
We employ a reverse cosine learning rate schedule during unlearning, in contrast to the commonly used cosine annealing strategy in standard training. 
Specifically, we define the schedule as follows:
\begin{align}
    \lambda_t = \lambda_\mathrm{min} + \frac{1}{2}(\lambda_\mathrm{max}-\lambda_\mathrm{min})\left(1-\cos\left(\frac{t}{T}\pi\right) \right),
\end{align}
where $\lambda_\mathrm{min}$ and $\lambda_\mathrm{max}$ denote the minimum and maximum learning rates, $T$ is the maximum number of epochs, and $t$ is the current epoch.
Although we report results up to 50 epochs, we set $T = 200$ to control the pace of learning rate increase in the early stages of unlearning.
We used the following learning rate settings for each LLM.
Unless otherwise noted, the same schedule was applied to both Ours (+GA) and Ours (+NPO) variants.
\begin{itemize}
  \item \textbf{Mistral-7B}: $(\lambda_\mathrm{min}, \lambda_\mathrm{max}) = (4\times10^{-7},\ 4\times10^{-5})$ for both \textsc{Wiki-Fact} and \textsc{TOFU}.
  \item \textbf{Llama3.1-8B}: $(1\times10^{-6},\ 1\times10^{-4})$ for both \textsc{Wiki-Fact} and \textsc{TOFU}.
  \item \textbf{Qwen2.5-7B}: $(1\times10^{-6},\ 1\times10^{-4})$ (evaluated only on \textsc{Wiki-Fact}).
  \item \textbf{Gemma2-9B}: $(1\times10^{-6},\ 1\times10^{-4})$ (evaluated only on \textsc{Wiki-Fact}).
\end{itemize}
For the CU metrics (NodeAcc and EdgeAcc), we restrict the evaluation to samples that the pre-unlearned LLM correctly answered, ensuring that we measure the extent to which memorized knowledge is forgotten.  
Under this setting, the pre-unlearned LLM achieves 100\% accuracy by definition.  
In contrast, we do not apply such normalization to the general utility score, as it is intended to assess the LLM’s overall performance on open-domain tasks.
To determine when forgetting has been successfully achieved, we perform a grid search over the number of unlearning epochs for each target entity.
Specifically, we evaluate NodeAcc and EdgeAcc every 5 epochs from 5 to 50, and record the results at the earliest epoch where both scores reach 0.
This early stopping scheme allows us to fairly compare forgetting efficiency across different methods while avoiding excessive updates.
For the ICU baseline, we use manually constructed prompts designed to suppress the target concept via in-context instruction. 
The prompt format is adapted from the original ICU setup proposed in the related paper~\citep{jin2024rwku}, and our specific implementation is shown in Example~\ref{icu_template}.

\begin{figure*}[tb]
  \centering
  \begin{minipage}{0.96\textwidth}
    \hrule
    \vspace{2pt}
    \noindent\textbf{Listing \thefigure:} Template of in-context unlearning (ICU).
    \vspace{2pt}
    \hrule
    \vspace{4pt}
\begin{lstlisting}[style=aaai-list]
You are an AI Assistant who is supposed to unlearn about Harry Potter and provide 
answers without its knowledge as if you never knew about it.
Don't tell anyone that you unlearned anything.
\end{lstlisting}
    \hrule
  \end{minipage}
  \caption{Template of in-context unlearning (ICU).}
  \label{icu_template}
\end{figure*}

\subsection{Datasets Details}
\label{dataset}
\paragraph{\textsc{Wiki-Fact} Dataset.}
The \textsc{Wiki-Fact} dataset is a real-world custom dataset of 60 human entities constructed for CU evaluation.  
Each entry in the dataset consists of the following four components:
\begin{itemize}
    \item \textbf{subject:} The entity's name, serving as the target entity for unlearning.
    \item \textbf{sentences:} The full Wikipedia article corresponding to the subject, segmented by paragraph. All mentions of the target entity are replaced with \texttt{[MASK]} tokens to facilitate evaluation of entity recall.
    \item \textbf{subj\_id:} A unique identifier linking the entity to its corresponding Wikidata entry.
    \item \textbf{attributions:} A set of knowledge triplets extracted from Wikidata, representing a collection of factual knowledge triplets about the subject (e.g., occupation, nationality, date of birth). These attributes are filtered using relation types defined in the \textsc{ParaRel} dataset~\citep{elazar-etal-2021-measuring}.
    Please see later in this section for more information on the \textsc{ParaRel} dataset.
\end{itemize}
This dataset is similar to the \textsc{T-REx} dataset~\citep{elsahar-etal-2018-rex}, which contains Wikipedia abstracts and associated attributions from Wikidata.
However, \textsc{T-REx} provides only short context (typically single-sentence abstracts) for each triplet, which limits its applicability for evaluating NU under naturalistic and information-rich conditions.
In contrast, \textsc{Wiki-Fact} includes the entire Wikipedia article per subject, enabling more realistic and contextually grounded evaluations, particularly important for NU, where the model is expected to suppress entity recall even when provided with multiple indirect clues (i.e., attributions) within a rich paragraph-level context.
An example entry for the subject \textit{Gilad Atzmon} in the \textsc{Wiki-Fact} is shown in Example~\ref{example_wf}. 
\paragraph{\textsc{TOFU} Dataset.}
The \textsc{TOFU} dataset~\citep{maini2024tofu} is a synthetic benchmark designed to evaluate unlearning performance in LLMs.
It consists of 200 fictional entities, each associated with 20 question-answer pairs.
For fine-tuning, we follow the original \textsc{TOFU} protocol, using instruction tuning with the 20 question–answer pairs per entity.
This ensures that all relevant knowledge is injected into the LLM in a controlled manner.
To evaluate NU, we construct five explanatory sentences per entity by combining content from four answer sentences.
These sentences provide rich, multi-fact contexts, allowing us to measure the LLM’s ability to suppress the target entity even when indirectly prompted.
To evaluate EU, we extract 20 structured knowledge triplets per entity by prompting GPT-4 with each question–answer pair.
These triplets, formatted as $(\bm{s}, \bm{r}, \bm{o})$, represent atomic facts and are used to assess whether the LLM forgets specific attributes of the target entity.
Each entry in our customized \textsc{TOFU} dataset contains:
\begin{itemize}
    \item \textbf{subject:} A unique fictional name used as the forgetting target.
    \item \textbf{sentences:} A short passage describing the entity, containing multiple factual statements (e.g., name, birthplace, affiliation).
    \item \textbf{attributions:} A list of structured $(\bm{s}, \bm{r}, \bm{o})$ knowledge triplets that summarize the facts in the sentence.
\end{itemize}
Unlike \textsc{Wiki-Fact}, which focuses on real-world individuals with rich and potentially redundant information, \textsc{TOFU} provides a minimal and synthetic evaluation setting that isolates memorization behavior in LLMs.
Importantly, since \textsc{TOFU} entities are fictional and not part of any LLM's pretraining corpus, the knowledge to be unlearned is entirely contained within the dataset itself.
This guarantees that unlearning performance can be assessed in a controlled and interpretable manner, without confounding effects from prior exposure or background knowledge.
\paragraph{\textsc{ParaRel} Dataset.}
The \textsc{ParaRel} dataset is a manually curated resource designed to probe factual knowledge in LLMs through cloze-style paraphrased prompts~\citep{elazar-etal-2021-measuring}.
It consists of 328 paraphrase patterns spanning 38 relations, such as \textit{place-of-birth}, \textit{occupation}, and \textit{works-for}.
Each relation is associated with multiple textual patterns that express the same factual relation between a subject and an object, enabling consistency evaluation across different surface forms.
\textsc{ParaRel} was constructed by extending the relation templates provided in LAMA~\citep{Petroni2019LanguageMA} and LPAQA~\citep{jiang-etal-2020-know}, followed by manual verification to ensure paraphrase correctness.
In addition, additional paraphrases were collected from Wikipedia sentences using syntax-aware search and further expanded through expert annotation.
An example for the \textit{place-of-birth} relation includes the patterns:
\begin{itemize}
    \item \texttt{[X] was born in [Y].}
    \item \texttt{[X] is a native of [Y].}
    \item \texttt{[X], born in [Y].}
\end{itemize}
In our work, \textsc{ParaRel} plays two key roles:
\begin{itemize}
\item For EU evaluation, we utilize its relation templates to convert knowledge triplets into natural language sentences, enabling fine-grained assessment of the LLM's ability to suppress specific factual associations.
\item In the Rel-Aware variant of the \textsc{Get\_Attr} function, we use its predefined relation types as guidance to steer the LLM toward generating a diverse and structured set of triplets during unlearning.
\end{itemize}

\begin{figure*}[tb]
  \centering
  \begin{minipage}{0.96\textwidth}
    \hrule
    \vspace{2pt}
    \noindent\textbf{Listing \thefigure:} Example from the \textsc{Wiki\_Fact} dataset.
    \vspace{2pt}
    \hrule
    \vspace{4pt}
\begin{lstlisting}[style=aaai-list]
[
  {
    "subject": "Gilad Atzmon",
    "sentences": [
      "[MASK] first became interested in British jazz when he came across recordings of Ronnie Scott and Tubby Hayes.[4] During his incapacitation for nearly a year following a climbing accident, [MASK] started playing the saxophone in earnest.[3] Discovering bebop, he said that the albums Charlie Parker with Strings were what made him want to be a jazz musician.[5]",
      "[MASK] has written satirical novels, non-fiction works and read essays on the subjects of Palestinian rights, Israel and identity politics. These writings have been described by scholars and anti-racism activists as being antisemitic and containing Holocaust denial.",
      ...
    ],
    "subj_id": "Q1389497",
    "attributions": [
      ["Gilad Atzmon", "place of birth", "Tel Aviv"],
      ["Gilad Atzmon", "field of work", "music"],
      ...
    ]
  },
  ...
]
\end{lstlisting}
    \hrule
  \end{minipage}
  \caption{Example from the \textsc{Wiki\_Fact} dataset.}
  \label{example_wf}
\end{figure*}

\section{Implementation Details of Our Proposal}
\label{implementation}
This section provides details on the further implementation of our proposed unlearning method.
We first describe the \textsc{Get\_Attr} function, which is central in extracting knowledge triplets from the LLM for use in the triplet-based loss.
Then, we introduce a relation-aware variant of this function that incorporates predefined relation types to improve coverage (Rel-Aware).
Finally, we describe the in-context prompting strategy used in \textsc{Get\_Attr} function to extract structured triplet format responses from the LLM.
\subsection{\textsc{Get\_Attr} function}
\label{get_attr}
In Algorithm~\ref {proposal_alg}, we use the \textsc{Get\_Attr} function to extract the triplets about the forgetting target from the LLM.
It consists of three processes: (1) a triplet extraction process that prompts the LLM to generate triplets about the forgetting target, (2) a triplet validation process that filters out incorrect or low-confidence knowledge by verifying each generated triplet using the pre-unlearned LLM, and (3) a triplet conversion process that transforms the validated triplets into natural language sentences for use in the unlearning loss.
\paragraph{Triplet Extraction Process}
In the triplet extraction process, we prompt the target LLM with $\bm{x}_{\mathrm{pro}}=$ ``Tell me about \{$\bm{e}^\mathrm{t}$\} in knowledge triplet format.'' and obtain a list of triplets related to the forgetting target.
We also use in-context learning, with examples shown in Exapmle~\ref{kg_extract}, to help the LLM output structured knowledge triplets.
Furthermore, to enforce the triplet format and ensure that each generated triplet centers around the forgetting target, we implement a decoding-time mechanism: at each newline character in the output, we automatically append the string ``($\bm{e}^\mathrm{t}$, '' to the beggining of the following line.
This technique forces the LLM to repeatedly generate knowledge triplets about $\bm{e}^\mathrm{t}$, each time starting with $\bm{e}^\mathrm{t}$ as the subject.
This subject-centric design simplifies the prompt format and reflects the common structure of entity-centric knowledge.
However, it also means that triplets where $\bm{e}^\mathrm{t}$ appears as an object are not extracted.
Since this procedure forces the LLM to generate knowledge related to $\bm{e}^\mathrm{t}$ regardless of confidence, it can include incorrect or hallucinated facts.
To address this, we apply a subsequent validation step that filters out such low-quality triplets based on correctness.
\paragraph{Triplet Validation Process}
In the triplet validation process, the candidate triplets generated during the triplet extraction process are evaluated for factual correctness using the pre-unlearned LLM (i.e., the LLM before applying any forgetting updates).
Each triplet $(\bm{e}^\mathrm{t}, \bm{r}, \bm{o})$ is individually assessed by constructing a verification prompt of the form: $\bm{x}_\mathrm{check}=$ ``Please check if the following knowledge is true or not. If the knowledge is true, say 1, and if not, say 0. ($\bm{e}^\mathrm{t}$, $\bm{r}$, $\bm{o}$)''
To improve reliability, we apply in-context learning by providing examples of clearly correct and incorrect triplets along with appropriate LLM responses, with examples shown in Example~\ref{kg_validate}.
The prompt is designed such that the LLM is expected to answer either ``1'' or ``0''.
Only the triplets for which the LLM responds with ``1'' are retained; the others are discarded as potentially incorrect or hallucinated.
This filtering step ensures that only triplets deemed reliable by the pre-unlearned LLM are used in subsequent unlearning, improving the precision of the forgetting process. \par
\paragraph{Triplet Convert Process}
After the triplets are validated, we convert each of them into natural language sentences to obtain the unlearning target sentences.
We leverage the pre-unlearned LLM to perform this conversion by prompting it with a format such as: ``Please convert the following knowledge triplet into ten natural texts. Knowledge: ($\bm{e}^\mathrm{t}$, $\bm{r}$, $\bm{o}$), Sentence: ''.
This allows us to obtain 10 diverse natural language sentences for each triplet, capturing the same underlying factual knowledge in multiple paraphrases.
To improve consistency and output quality, we apply in-context learning using examples shown in Example~\ref{kg_convert}.
Each generated sentence is then split into two parts: $\bm{X}_\mathrm{ent}$, which includes the text up to the target entity token $\bm{e}^\mathrm{t}$, and $\bm{X}_\mathrm{attr}$, which includes the remaining portion describing the attribution (e.g., $\bm{X}_\mathrm{ent}=$ Harry Potter'', $\bm{X}_\mathrm{attr}=$ defeated Voldemort.'').
This split allows the unlearning process to focus on the attributional tokens. \par
Together, these three processes—triplet extraction, triplet validation, and triplet conversion—compose the full implementation of the \textsc{Get\_Attr} function.
This function plays a central role in extracting high-quality, structured, and verifiable representations of knowledge related to the forgetting target.
The complete procedure is outlined in Algorithm~\ref{getattr_alg}.

\begin{figure}[tb]
\begin{algorithm}[H]
\caption{\textsc{Get\_Attr} algorithm}
\label{getattr_alg}
\begin{algorithmic}[1]
\REQUIRE Forgetting target: $\bm{e}^\mathrm{t}$, Target LLM: $\bm{\theta}$, Pre-unlearned LLM: $\bm{\theta}_\mathrm{pre}$, Generation steps: $T$
\STATE $\bm{x}_{\mathrm{prompt}} \leftarrow$ ``Tell me about \{$\bm{e}^\mathrm{t}$\} in the knowledge triplet format.''
\STATE $\bm{y}_{\mathrm{out}} \leftarrow$ null
\FOR{$t = 1$ to $T$}
    \STATE $\bm{y}_{\mathrm{tmp}} \leftarrow \arg\max_y p_{\bm{\theta}} (y|[\bm{x}_{\mathrm{prompt}}, \bm{y}_{\mathrm{out}}])$
    \STATE $\bm{y}_{\mathrm{out}} \leftarrow \bm{y}_{\mathrm{out}} + \bm{y}_{\mathrm{tmp}}$
    \IF{$\bm{y}_{\mathrm{tmp}}$ is ``\textbackslash n''}
        \STATE // Force continuation in triplet format by inserting entity token.
        \STATE $\bm{y}_{\mathrm{out}} \leftarrow \bm{y}_{\mathrm{out}} +$ ``(\{$\bm{e}^\mathrm{t}$\}, ''
    \ENDIF
\ENDFOR
\STATE $\mathcal{T} \leftarrow$ Split $\bm{y}_{\mathrm{out}}$ by newline characters
\STATE $\bm{X}_{\mathrm{ent}}, \bm{X}_{\mathrm{attr}} \leftarrow \emptyset, \emptyset$
\FOR{each triplet $t_i \in \mathcal{T}$}
    \STATE // Triplet validation using pre-unlearned LLM
    \STATE $\bm{x}_{\mathrm{check}} \leftarrow$ ``Please check if the following knowledge is true or not. If the knowledge is true, say 1, and if not, say 0. ($\bm{e}^\mathrm{t}$, $\bm{r}$, $\bm{o}$)''
    \STATE $\bm{r}_i \leftarrow \mathrm{argmax}_yp_{\bm{\theta}_\mathrm{pre}}(y|\bm{x}_{\mathrm{check}})$
    \IF{$\bm{r}_i$ == ``1''}
        \STATE // Convert triplet to sentence using pre-unlearned LLM
        \STATE $\bm{x}_{\mathrm{convert}} \leftarrow$ ``Please convert the following knowledge triplet into a natural sentence: \{$t_i$\}''
        \STATE $\bm{x}_i \leftarrow \mathrm{argmax}_{\bm{y}}p_{\bm{\theta}_\mathrm{pre}}(\bm{y}|\bm{x}_{\mathrm{convert}})$
        \STATE // Split sentence at the entity mention
        \STATE Find index $k$ of $\bm{e}^\mathrm{t}$ in $\bm{x}_i$
        \STATE $\bm{x}_{\mathrm{ent}} \leftarrow \bm{x}_i[:k+\texttt{len}(\bm{e}^\mathrm{t})]$
        \STATE $\bm{x}_{\mathrm{attr}} \leftarrow \bm{x}_i[k+\texttt{len}(\bm{e}^\mathrm{t}):]$
        \STATE $\bm{X}_{\mathrm{ent}} \leftarrow \bm{X}_{\mathrm{ent}} \cup \{\bm{x}_{\mathrm{ent}}\}$
        \STATE $\bm{X}_{\mathrm{attr}} \leftarrow \bm{X}_{\mathrm{attr}} \cup \{\bm{x}_{\mathrm{attr}}\}$
    \ENDIF
\ENDFOR
\RETURN $\bm{X}_\mathrm{ent}$, $\bm{X}_\mathrm{attr}$
\end{algorithmic}
\end{algorithm}
\end{figure}

\subsection{\textsc{Get\_Attr} with Relation Types}
\label{get_attr_rel}
To enhance the coverage of the knowledge graph captured during the unlearning process (KG coverage), we introduce a relation-aware variant of the \textsc{Get\_Attr} function (Rel-Aware).
This variant enriches the triplet extraction prompt by explicitly specifying predefined relation types, encouraging the LLM to generate diverse facts about the forgetting target.
Concretely, in line 8 of Algorithm~\ref{getattr_alg}, where the standard method appends (\{$\bm{e}^\mathrm{t}$\}, '', the Rel-Aware variant instead iterates over a list of relation types $\bm{r}$ and constructs prompts in the form ({$\bm{e}^\mathrm{t}$}, $\bm{r}$, '', thereby steering the LLM toward relation-specific completions. 
We use the relation types defined by the \textsc{ParaRel} dataset as guidance (see Sec.~\ref{dataset} for dataset details), applying them sequentially regardless of the pre-unlearned LLM’s ability to answer them correctly.
This procedure allows us to cover both well-known and less salient relations, promoting more comprehensive and diverse triplet extraction and enhancing KG coverage.

\begin{figure*}[tb]
  \centering
  \begin{minipage}{0.96\textwidth}
    \hrule
    \vspace{2pt}
    \noindent\textbf{Listing \thefigure:} In-context prompt of triplet extraction process.
    \vspace{2pt}
    \hrule
    \vspace{4pt}
\begin{lstlisting}[style=aaai-list]
Knowledge: (Finland, capital, Helsinki)
Subject: Mary I of England
Knowledge:
(Mary I of England, father, Henry VIII)
(Mary I of England, follows, Lady Jane Grey)
(Mary I of England, sibling, Elizabeth I)
(Mary I of England, sibling, Edward VI)
(Mary I of England, followed by, Elizabeth I)
(Mary I of England, country of citizenship, England)
(Mary I of England, religion or worldview, Roman Catholicism)
 
Subject: My Neighbor Totoro
Knowledge:
(My Neighbor Totoro, production company, Studio Ghibli)
(My Neighbor Totoro, screenwriter, Hayao Miyazaki)
(My Neighbor Totoro, director, Hayao Miyazaki)
(My Neighbor Totoro, cast member, Noriko Hidaka)
(My Neighbor Totoro, country of origin, Japan)
(My Neighbor Totoro, genre, fantasy film)
(My Neighbor Totoro, cast member, Chika Sakamoto)
\end{lstlisting}
    \hrule
  \end{minipage}
  \caption{In-context prompt of triplet extraction process.}
  \label{kg_extract}
\end{figure*}

\begin{figure*}[tb]
  \centering
  \begin{minipage}{0.96\textwidth}
    \hrule
    \vspace{2pt}
    \noindent\textbf{Listing \thefigure:} In-context prompt of triplet validation process.
    \vspace{2pt}
    \hrule
    \vspace{4pt}
\begin{lstlisting}[style=aaai-list]
Knowledge: (Finland, capital, Helsinki)
Answer: 1

Knowledge: (My Neighbor Totoro, director, Steven)
Answer: 0

Knowledge: (Mary I of England, father, Henry VIII)
Answer: 1
\end{lstlisting}
    \hrule
  \end{minipage}
  \caption{In-context prompt of triplet validation process.}
  \label{kg_validate}
\end{figure*}

\begin{figure*}[tb]
  \centering
  \begin{minipage}{0.96\textwidth}
    \hrule
    \vspace{2pt}
    \noindent\textbf{Listing \thefigure:} In-context prompt for triplet-to-sentence conversion.
    \vspace{2pt}
    \hrule
    \vspace{4pt}
\begin{lstlisting}[style=aaai-list]
Knowledge: (Mary I of England, father, Henry VIII) 
Sentence: 
The father of Mary I of England is Henry VIII 
Who is the father of Mary I of England? Henry VIII 
Mary I of England's father is Henry VIII
Mary I of England was the daughter of Henry VIII 
The King of England and father of Mary I was Henry VIII 
One of the Tudor monarchs, who fathered Mary I, was Henry VIII
The father of Queen Mary I, a significant figure 
in English history, was Henry VIII 
Mary I's father, a famous Tudor king, was Henry VIII
The English ruler who was Mary I's father was Henry VIII 
Queen Mary I was born to the King of England, Henry VIII 

Knowledge: (My Neighbor Totoro, director, Hayao Miyazaki) 
Sentence: 
The director of My Neighbor Totoro is Hayao Miyazaki 
Who is the director of My Neighbor Totoro? Hayao Miyazaki 
My Neighbor Totoro was directed by Hayao Miyazaki 
The person who directed My Neighbor Totoro was Hayao Miyazaki 
The filmmaker behind My Neighbor Totoro was Hayao Miyazaki 
My Neighbor Totoro was created under the direction of Hayao Miyazaki 
The animation director responsible for My Neighbor Totoro
was Hayao Miyazaki 
The visionary director of My Neighbor Totoro was Hayao Miyazaki 
My Neighbor Totoro was brought to life by the direction
of Hayao Miyazaki 
The Studio Ghibli director who worked on My Neighbor Totoro
was Hayao Miyazaki 

Knowledge: (Finland, capital, Helsinki)
Sentence: 
The capital of Finland is Helsinki 
Where is the capital of Finland? Helsinki
Which city is the capital of Finland? Helsinki 
The city that serves as the capital of Finland is Helsinki 
The administrative center of Finland is Helsinki 
Finland's most important capital city is Helsinki 
The capital and largest city of Finland is Helsinki
The government of Finland is based in Helsinki
The political and cultural hub of Finland is Helsinki 
The Finnish capital city is Helsinki 
\end{lstlisting}
    \hrule
  \end{minipage}
  \caption{In-context prompt for triplet-to-sentence conversion.}
  \label{kg_convert}
\end{figure*}

\end{document}